\documentclass[a4paper,fleqn]{cas-sc}
\ifdefined\pdfobjcompresslevel\pdfobjcompresslevel=0\fi
\usepackage{hyperref}
\usepackage[numbers,sort&compress]{natbib}
\usepackage{graphicx}
\graphicspath{{./}}
\usepackage{textcomp}
\usepackage{float}       
\usepackage{adjustbox}   
\usepackage{url}
\hypersetup{citecolor=blue,linkcolor=blue,urlcolor=blue,breaklinks=true}

\makeatletter
\ExplSyntaxOn
\RenewDocumentEnvironment { figure } { O{} }
  {
    \__reset_fig:
    \bool_gset_false:N \g_fig_full_bool
    \dim_set:Nn \l_fig_width_dim { \linewidth }
    \keys_set:nn { cas / fig } { #1 }
    \csxdef{fps@figure}{\l_fig_pos_tl}
    \csgdef{figwidth}{\dim_use:N \l_fig_width_dim}
    \cs_set_eq:NN \@makecaption \__make_fig_caption:nn
    \exp_last_unbraced:Nx \@float { {figure} [ \l_fig_pos_tl ] }
     \l_fig_align_tl
     \sffamily\small
  }
  {
    \tl_if_eq:NnTF \l_fig_pos_tl { H } { \float@endH } { \end@float }
  }
\ExplSyntaxOff
\makeatother

\newcommand{\mmAcc}{0.763}
\newcommand{\mmAccS}{0.756}
\newcommand{\mmMf}{0.664}
\newcommand{\mmK}{0.656}
\newcommand{\mmAuc}{0.825}
\newcommand{\mmAp}{0.487}
\newcommand{\mmRho}{0.765}

\begin{document}
\let\WriteBookmarks\relax
\def\floatpagepagefraction{1}
\def\textpagefraction{.001}

\shorttitle{MM-Net: sleep staging and sleep apnea assessment after ischemic stroke}
\shortauthors{M. W. Suva et~al.}

\title[mode = title]{MM-Net: A Multimodal Multi-Task Network for Sleep Staging and Sleep Apnea Assessment After Ischemic Stroke}

\author[1]{Md Wahiduzzaman Suva}[orcid=0009-0007-6227-7282]
\ead{26-94088-2@student.aiub.edu}

\author[1]{Esm E Moula Chowdhury Abha}[orcid=0009-0008-3776-2283]
\ead{26-94089-2@student.aiub.edu}

\author[1]{Md Imtiaj Alam Sajin}[orcid=0009-0009-2423-1835]
\ead{26-94090-2@student.aiub.edu}

\author[1]{Md Iftekharul Mobin}[orcid=0000-0002-3065-2486]
\cormark[1]
\ead{iftekhar.mobin@aiub.edu}

\author[4]{Jia Uddin}[orcid=0000-0002-3403-4095]
\ead{jia.uddin@wsu.ac.kr}

\affiliation[1]{organization={Department of Computer Science and Engineering, American International University-Bangladesh},
    city={Dhaka},
    postcode={1229},
    country={Bangladesh}}

\affiliation[4]{organization={AI and Big Data Department, Woosong University},
    city={Daejeon},
    country={Republic of Korea}}

\cortext[1]{Corresponding authors: M. I. Mobin (iftekhar.mobin@aiub.edu)}

\begin{abstract}
Automated polysomnography after ischemic stroke presents challenges arising from altered sleep-related neural activity and concurrent respiratory disturbances. Existing stroke-specific approaches have primarily addressed sleep-stage classification, leaving the joint analysis of sleep architecture and respiratory events insufficiently investigated. Furthermore, adapting pretrained EEG encoders introduces a representation inconsistency when downstream models are trained on embeddings from previously observed patients but evaluated on unseen individuals. We propose MM-Net, a multimodal multi-task framework that addresses this inconsistency through cross-fitting and ridge-based embedding alignment. Sleep-adapted EEG--EOG representations and engineered neural descriptors are integrated with learned cardiorespiratory features through cross-attention and shared temporal modeling to support simultaneous sleep staging and respiratory-event detection. On iSLEEPS, comprising 99 patients with subacute ischemic stroke, ten-fold patient-independent evaluation achieved 76.3\% sleep-staging accuracy (Cohen's $\kappa = 0.656$) and 0.825 respiratory-event AUC, outperforming the evaluated baselines under matched protocols. Controlled experiments showed that the sleep-adapted encoder improves the downstream model when its embeddings are aligned, raising staging accuracy by 1.4 percentage points over the frozen encoder ($p = 0.020$), against 0.2 points without alignment. The framework also achieved 86.4\% staging accuracy on healthy Sleep-EDF, above every single-EEG architecture reimplemented on the same recordings, and 80.3\% staging accuracy with 0.889 respiratory AUC on ISRUC-I, exceeding its retrained staging baselines. Without retraining, the iSLEEPS-trained model retained respiratory AUCs of 0.833 on ISRUC-I and 0.841 on UCD. These findings support representation-consistent neural adaptation and joint physiological modeling across stroke, healthy-sleep, and sleep-disorder cohorts.
\end{abstract}

\begin{keywords}
Sleep staging \sep sleep apnea \sep ischemic stroke \sep polysomnography \sep EEG representation learning \sep multi-task learning
\end{keywords}

\maketitle

\section{Introduction}

Sleep disturbances are common after ischemic stroke and have been associated with neurological recovery and functional outcomes~\cite{bassetti2006sleep,denis2024sleep,tellenbach2023brainstem,johnson2010frequency}. Sleep-disordered breathing is particularly prevalent in this population, making the assessment of sleep architecture and respiratory disturbance relevant to post-stroke evaluation. Polysomnography (PSG) records the physiological information required for both assessments, including electroencephalographic (EEG), electrooculographic (EOG), electromyographic (EMG), respiratory, and cardiovascular signals. Sleep-stage scoring primarily depends on neural and muscular activity, whereas respiratory-event identification relies on changes in ventilation, respiratory effort, oxygen saturation, and related physiological responses. Their concurrent analysis therefore requires a model capable of learning from heterogeneous signals whose relevance differs across prediction tasks.

Automated sleep staging has progressed from convolutional and recurrent architectures to attention-based sequence models and pretrained neural representations~\cite{supratak2017deepsleepnet,phan2019seqsleepnet,eldele2021attnsleep,supratak2020tinysleepnet,labram2024,cbramod2025,eegpt2024,biot2023}. Multimodal approaches have expanded this direction by integrating physiological signals, temporal context, and complementary representations. Recent studies, including MMS-SleepNet~\cite{mmssleepnet2025} and CrossFusionSleepNet~\cite{crossfusionsleepnet2026}, demonstrate the growing use of physiological knowledge, modality-specific encoding, and attention-based fusion for sleep staging. Respiratory-event detection has developed alongside these methods, and joint approaches have established the feasibility of predicting sleep stages and respiratory events within a shared framework. Huttunen \emph{et al.}~\cite{huttunen2023simultaneous} evaluated simultaneous scoring in 877 clinical PSG recordings, while subsequent multi-task models have investigated related problems in other sleep populations~\cite{multitask2025elderly}. These developments establish a methodological foundation for joint sleep analysis, although their evaluation has primarily involved general sleep-disorder, suspected-apnea, or elderly cohorts.

The application of these methods to ischemic stroke introduces additional physiological and computational considerations. Cerebrovascular injury can alter sleep spindles, slow-wave activity, and hemispheric EEG characteristics~\cite{gottselig2002stroke,fernandez2019spindles,facchin2020slowwaves,simpson2023laterality,zheng2025spindles,anastasi2017bsi}, potentially affecting representations learned from healthy or conventional sleep-disorder populations. The recently released iSLEEPS dataset~\cite{maiti2026isleeps} provides multimodal overnight PSG recordings from 100 patients with ischemic stroke, including expert annotations of sleep stages and respiratory events. Its initial analysis established sleep-staging baselines for this population. Bose \emph{et al.}~\cite{bose2026gap} subsequently investigated the generalization of EEG-based staging models between healthy and stroke cohorts, reporting performance degradation and changes in the physiological patterns emphasized by the model. More recently, Alhazmi~\cite{alhazmi2026sleepstagenet} introduced SleepStageNet, a compact five-channel staging architecture evaluated on 95 iSLEEPS recordings, together with physiological attribution analyses and transfer to an independent healthy-sleep cohort. Collectively, these studies have advanced stroke-specific sleep classification and demonstrated the importance of population-dependent evaluation. The joint prediction of sleep architecture and respiratory-event onset from stroke PSG, however, remains insufficiently investigated.

This gap also raises a question concerning the organization of multimodal physiological information. Neural and cardiorespiratory signals provide different evidence for sleep state and respiratory disturbance, although both tasks depend on temporal context and interactions among physiological systems. A shared model must therefore integrate information across modalities while retaining features relevant to individual outputs. Furthermore, respiratory-event annotations are closely associated with recorded airflow and respiratory-effort signals. Improved discrimination from multimodal inputs cannot, by itself, establish whether a model has learned complementary physiological relationships or relies predominantly on channels directly associated with event scoring. Controlled modality and channel interventions are necessary to distinguish these contributions and characterize the dependence of each task on its available signals.

A separate methodological difficulty arises when pretrained EEG encoders are adapted to a clinical cohort and subsequently used to train a downstream prediction model. Adaptation can improve the relevance of learned representations to post-stroke sleep, but embeddings generated for patients used during encoder training may differ from those produced for previously unseen patients. Cross-fitting addresses the in-sample dependence by generating downstream training embeddings from encoders that have not observed the corresponding patients, consistent with the cross-fitting principle of double machine learning~\cite{chernozhukov2018dml}. However, independently adapted encoders may produce representations in different feature spaces, introducing an additional inconsistency with the encoder used at deployment. Consequently, reliable integration requires both out-of-sample representation generation and alignment with the deployed encoder's embedding space. The contribution of these operations can be evaluated directly through controlled encoder-integration comparisons.

To address these problems, we develop MM-Net, a multimodal multi-task framework for joint sleep-stage classification and respiratory-event-onset detection. The neural stream combines physiologically defined descriptors with embeddings from a sleep-adapted LaBraM encoder~\cite{labram2024}. Cross-fitting produces out-of-sample representations for downstream training, and ridge-based alignment maps these representations into the embedding space of the encoder used for inference. A complementary convolutional stream extracts information directly from cardiorespiratory waveforms. Cross-attention and bidirectional temporal modeling integrate the representations, while task-specific prediction pathways support five-class sleep staging and respiratory-event scoring. A direct cardiorespiratory pathway preserves respiratory information for event detection, and the resulting scores additionally support patient-level apnea--hypopnea index estimation.

We evaluate the framework using patient-independent protocols across stroke, healthy-sleep, and sleep-disorder populations. The primary analysis uses 99 iSLEEPS patients, supported by matched staging and respiratory baselines, controlled encoder comparisons, and physiological signal interventions. Controlled experiments further establish the contribution of embedding alignment to sleep-adapted encoder integration. To examine applicability beyond the development population, the framework is separately evaluated on Sleep-EDF and ISRUC-Sleep. Additional experiments assess transfer of the iSLEEPS-trained model to independent cohorts without retraining. Together, these evaluations investigate both the methodological contribution of representation-consistent adaptation and the applicability of joint physiological modeling across different sleep populations.

The principal contributions of this study are threefold:

\begin{enumerate}

\item Representation-consistent integration of sleep-adapted EEG features. We address the mismatch arising when an EEG encoder adapted on training patients supplies representations to a downstream model evaluated on unseen patients. A cross-fitting and embedding-alignment procedure integrates out-of-sample neural representations into a common deployment space, enabling the downstream framework to benefit from sleep-specific adaptation without relying on in-sample encoder outputs during training.

\item Task-specialized multimodal modeling of post-stroke sleep physiology. We develop a unified framework for sleep-stage classification and respiratory-event detection that combines neural and cardiorespiratory representations through shared temporal modeling while retaining task-specific respiratory information. This design enables concurrent characterization of sleep architecture and respiratory disturbances in subacute ischemic stroke, with patient-level AHI estimation as an additional output.

\item Evidence for physiological specialization and cross-cohort applicability. Through controlled signal interventions and patient-independent evaluations, we examine the dependence of each prediction task on its physiological inputs and establish the performance of the integrated framework against matched baselines. Independent-cohort experiments further assess the transfer of the learned representations beyond the stroke population.

\end{enumerate}

The remainder of this paper presents the related literature, datasets, proposed methodology, experimental evaluation, and discussion of the findings.

\section{Related Work}
\label{sec:related}

\subsection{Automatic Sleep Staging}

Automatic sleep staging has progressed from feature-based classifiers to deep
models that learn temporal and spectral representations directly from PSG
signals. DeepSleepNet combines convolutional feature extraction with recurrent sequence modeling~\cite{supratak2017deepsleepnet}, and SeqSleepNet formulates sleep staging as sequence-to-sequence classification across consecutive epochs~\cite{phan2019seqsleepnet}. AttnSleep uses multi-resolution convolution and
attention to capture complementary temporal characteristics
~\cite{eldele2021attnsleep}, and TinySleepNet reduces model complexity while
retaining recurrent modeling of stage transitions
~\cite{supratak2020tinysleepnet}. U-Time approaches staging as dense
time-series segmentation~\cite{perslev2019utime}, whereas XSleepNet combines
raw-signal and spectral representations within a common sequence model
~\cite{phan2021xsleepnet}.

More recent work has expanded the range of temporal models and learned
representations. SleepTransformer applies attention-based sequence modeling
~\cite{sleeptransformer2022}, L-SeqSleepNet extends modeling across longer
sleep-stage sequences~\cite{lseqsleepnet2023}, and ZleepAnlystNet and
FlexSleepTransformer investigate alternative training strategies and flexible
channel configurations
~\cite{jirakittayakorn2024zleep,guo2024flexsleep}. State-space architectures
based on Mamba have also been explored for sleep staging
~\cite{gu2023mamba,bitmamsleep2024}. In parallel, large pretrained EEG models
such as LaBraM, CBraMod, EEGPT, and BIOT provide representations learned from
substantially larger collections of neural recordings
~\cite{labram2024,cbramod2025,eegpt2024,biot2023}. Such encoders are typically
used frozen or fine-tuned end to end for a single task. Using a fine-tuned
encoder as the input of a second model trained on the same patients raises the
requirement addressed by cross-fitting, namely that the
second model be trained on out-of-sample first-stage outputs
~\cite{wolpert1992stacked,chernozhukov2018dml}.

Most of these developments have been established on healthy or general sleep
datasets. Performance can change substantially when models are transferred
across cohorts because acquisition protocols, channel configurations, patient
characteristics, and sleep physiology differ between datasets
~\cite{guillot2023transfer}. This issue is especially relevant after stroke,
where the neural patterns used in conventional sleep staging may themselves be
altered by neurological injury.

\subsection{Multimodal and Joint PSG Analysis}

Clinical PSG contains several complementary physiological modalities, and a
parallel line of work has examined how these signals can be incorporated into
automatic sleep analysis. U-Sleep, for example, was developed to operate across
different EEG and EOG derivations and multiple sleep cohorts
~\cite{perslev2021usleep}. Other multimodal approaches combine neural, ocular,
muscular, cardiac, or respiratory signals through early fusion,
modality-specific encoders, or higher-level feature fusion. Recent work has also extended representation learning across PSG modalities, including SleepFM~\cite{sleepfm2026}. Cross-scale representations, self-supervised learning, and standardized sleep benchmarks have further broadened the range of models evaluated across datasets~\cite{csbrain2025,sleepyland2026,estevan2025ssl}.

Joint modeling of sleep stage and sleep-related disorders has received less
attention than sleep staging alone. Huttunen \emph{et al.}
~\cite{huttunen2023simultaneous} trained a single network for simultaneous
sleep-stage and respiratory-event scoring across 877 clinical recordings and
examined input configurations ranging from pulse oximetry to a broader PSG
montage. Multi-task approaches combining sleep staging with apnea-related
prediction have subsequently been investigated in other clinical populations
~\cite{multitask2025elderly}. These studies demonstrate that the two outputs can
be learned within a common model, but they have primarily been evaluated in
suspected-apnea, elderly, or general sleep populations.

\subsection{Respiratory-Event Detection}

Automated analysis of sleep-disordered breathing has largely developed as a
separate task. Apnea and hypopnea detectors commonly use airflow, thoracic and
abdominal respiratory effort, oxygen saturation, and ECG-derived cardiac
measures. These signals describe different stages of the physiological response:
airflow reflects reduction or cessation of ventilation, respiratory effort
provides information about continued or absent breathing effort, and oximetry
captures the subsequent change in blood oxygenation. Oxygen desaturation also
forms part of the clinical criteria used in hypopnea scoring
~\cite{aasm2023manual}. End-to-end deep models have also been developed for
respiratory-event scoring, including detection from a single respiratory-effort
belt~\cite{nassi2022effortbelt}. Biswal \emph{et al.}~\cite{biswal2018expert}
trained a recurrent--convolutional network to jointly score sleep stages, apnea
events, and limb movements from clinical PSG.

Respiratory-event models are typically optimized independently of sleep
staging and often operate on temporal resolutions different from the 30-s epoch
used for stage classification. Joint models reduce this separation by learning
both outputs from the same recording.

\subsection{Sleep and Sleep-Disordered Breathing After Stroke}

Sleep physiology is frequently altered after stroke, and sleep disturbances
have been associated with neurological and functional recovery
~\cite{bassetti2006sleep,denis2024sleep,tellenbach2023brainstem}. Changes in
sleep-related neural activity have also been reported after cerebrovascular
injury. Previous studies describe alterations in sleep spindles, slow-wave activity, and related electrophysiological patterns~\cite{gottselig2002stroke,fernandez2019spindles,simpson2023laterality,facchin2020slowwaves,zheng2025spindles}; quantitative EEG measures such as the brain symmetry index have also been examined in relation to hemispheric injury~\cite{anastasi2017bsi}. These changes make
stroke sleep a distinct setting for models whose representations were
developed predominantly on non-stroke populations.

Sleep-disordered breathing is also common during the subacute phase, so sleep
state and respiratory disturbance coexist within the same clinical recording.
Despite this overlap, automated stroke-specific sleep analysis remains limited.
The iSLEEPS dataset introduced a dedicated PSG resource for subacute ischemic
stroke and reported baseline sleep-staging performance
~\cite{maiti2026isleeps}; subsequent work has further examined the behavior of
deep staging models on this cohort~\cite{bose2026gap,alhazmi2026sleepstagenet}. Existing stroke-specific
studies have therefore concentrated primarily on sleep-stage classification.
Joint epoch-level modeling of sleep stage and respiratory disturbance remains
insufficiently characterized in this population.

\section{Datasets and Preprocessing}
\label{sec:data}

Four publicly available polysomnography datasets are used in this study (Table~\ref{tab:datasets}):
iSLEEPS~\cite{maiti2026isleeps}, Sleep-EDF
~\cite{kemp2000sleepedf,goldberger2000physionet}, ISRUC-Sleep
~\cite{khalighi2016isruc} with its three subgroups, and the St Vincent's
University Hospital / University College Dublin sleep apnea database (UCD)
~\cite{ucddb2007}. iSLEEPS is the primary cohort used for model development and
patient-independent evaluation. Sleep-EDF is used as a healthy-sleep reference
for implementation checks, pretraining of a transferred baseline, and
subject-independent evaluation of the staging model outside the stroke cohort. ISRUC-Sleep provides
independent sleep-disorder and healthy cohorts with both sleep-stage and
respiratory-event annotations; it is used both to train and evaluate MM-Net
under subject-independent cross-validation and to test the iSLEEPS-trained model
without retraining. UCD provides a further sleep-apnea cohort recorded on a
different acquisition system and is used for transfer only. No external cohort is
used to select the reported iSLEEPS model configuration.

\begin{table}[H]
\caption{Datasets used in the study and their roles in the evaluation.}
\label{tab:datasets}
\centering
\small
\setlength{\tabcolsep}{4pt}
\begin{tabular}{>{\raggedright\arraybackslash}p{0.21\columnwidth} >{\raggedright\arraybackslash}p{0.24\columnwidth}
                >{\raggedright\arraybackslash}p{0.18\columnwidth} >{\raggedright\arraybackslash}p{0.27\columnwidth}}
\toprule
Dataset & Population & Data used & Role \\
\midrule
iSLEEPS~\cite{maiti2026isleeps}
& Subacute ischemic stroke
& 99 patients
& Development and in-domain evaluation \\

Sleep-EDF~\cite{kemp2000sleepedf,goldberger2000physionet}
& Healthy sleep
& 40 recordings
& Reimplementation checks, healthy-sleep reference, and external staging \\

ISRUC-Sleep I~\cite{khalighi2016isruc}
& Sleep disorders
& 100 patients
& Cross-validation and transfer (staging, respiratory, AHI) \\

ISRUC-Sleep II~\cite{khalighi2016isruc}
& Sleep disorders
& 8 subjects, two nights each
& Cross-validation and transfer \\

ISRUC-Sleep III~\cite{khalighi2016isruc}
& Healthy
& 10 subjects
& Cross-validation and transfer \\

UCD~\cite{ucddb2007}
& Suspected sleep apnea
& 25 patients
& Transfer (staging, respiratory, AHI) \\
\bottomrule
\end{tabular}
\end{table}

\subsection{iSLEEPS: Primary Stroke Cohort}

The iSLEEPS dataset contains overnight PSG from 100 patients with subacute
ischemic stroke~\cite{maiti2026isleeps}. The released cohort has a mean age of
$50.5$ years and includes 77 male and 23 female participants. Sleep is scored
into the five AASM stages: Wake (W), N1, N2, N3, and rapid-eye-movement (REM)
sleep. Respiratory events are also annotated throughout the recordings.

The public release contains 100 recordings. SN15 and SN28 contain the same overnight
recording, so SN28 is excluded, leaving $N=99$ patients and $92{,}560$ 30-s epochs for the
analyses reported here.

\begin{figure}[pos=H]
\centering
\includegraphics[width=0.55\textwidth]{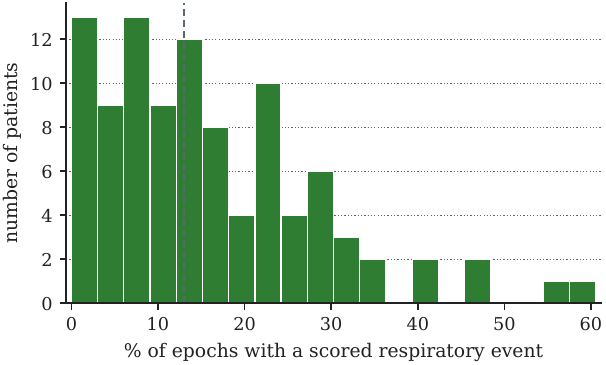}
\caption{Patient-level respiratory-event burden in iSLEEPS, measured as the
fraction of 30-s epochs containing the onset of a scored respiratory event. The dashed line
denotes the cohort median of $13\%$.}
\label{fig:prev}
\end{figure}

The resulting sleep-stage distribution is shown in
Table~\ref{tab:dist}. N2 accounts for the largest proportion of epochs, and N1 and N3 are the least frequent stages. Respiratory-event onsets are marked in $16.0\%$ of all
epochs. At the patient level, the fraction of event-positive epochs ranges from
$0$ to $60.5\%$; 80 of the 99 patients have respiratory events in more than
$5\%$ of their epochs, and the median burden is $13\%$
(Figure~\ref{fig:prev}).

\begin{table}[H]
\caption{Sleep-stage distribution in the analyzed iSLEEPS cohort
($N=99$; $92{,}560$ epochs).}
\label{tab:dist}
\centering
\begin{tabular}{lccccc}
\toprule
Stage & W & N1 & N2 & N3 & REM \\
\midrule
Share (\%) & 26.7 & 10.1 & 42.3 & 8.9 & 12.1 \\
\bottomrule
\end{tabular}
\end{table}

\subsection{Sleep-EDF: Healthy-Sleep Reference Cohort}

Sleep-EDF provides the healthy-sleep reference~\cite{kemp2000sleepedf,goldberger2000physionet}.
The experiments use the Sleep Cassette subset corresponding to subjects 0--20, comprising 40
overnight recordings, with splits constructed at the subject level because individual subjects
contribute more than one night. Sleep-EDF serves to check the reimplemented staging architectures
in the healthy-sleep setting in which they were originally evaluated, to pretrain the transferred
baseline, and to evaluate the proposed staging model on an independent healthy-sleep cohort. The
database provides the EEG derivations Fpz-Cz and Pz-Oz but not the cardiorespiratory signal set or
respiratory-event annotations, so respiratory performance is not evaluated on Sleep-EDF.

\subsection{ISRUC-Sleep: Independent Sleep-Disorder Cohort}

ISRUC-Sleep provides the independent evaluation in populations with and without
sleep disorders~\cite{khalighi2016isruc}. Subgroup I contains $100$ patients with
sleep disorders recorded on one night each, subgroup II eight patients recorded
on two nights each, and subgroup III ten healthy adults. The recordings include
central and occipital EEG derivations corresponding to C4, C3, O2, and O1
referenced to the contralateral mastoid, two EOG derivations and chin EMG,
together with oronasal airflow, nasal pressure, thoracic and abdominal effort,
SpO$_2$, and ECG, and annotations of obstructive, central, and mixed apneas and
hypopneas; the annotations of the first scorer are used. Summed effort is formed
as the sum of the thoracic and abdominal traces, and pulse as the beat-to-beat
heart rate derived from ECG R-peaks, so that every iSLEEPS input has an ISRUC
counterpart. In $30$ of the $126$ ISRUC recordings the EDF header declares the
EEG and EOG range as $\pm25\,\mu$V, which places their N3 peak-to-peak amplitude
at $9$--$23\,\mu$V against $98$--$283\,\mu$V in the remaining recordings and below
the $75\,\mu$V that defines N3; these channels are restored by a factor of ten, a
correction decided from the header alone. One subgroup-I recording stores
monopolar EEG and EOG, from which the mastoid-referenced derivations are formed.

ISRUC-Sleep therefore supports evaluation of every output of the proposed
framework: sleep-stage prediction, respiratory-event prediction and, in subgroup I,
AHI estimation across the full severity range.

\subsection{UCD: Independent Sleep-Apnea Cohort}

The UCD database~\cite{ucddb2007,goldberger2000physionet} contains overnight PSG
from $25$ adults referred for suspected sleep-disordered breathing, with annotated
AHI between $1.2$ and $81.2$ events/h. It records C3 and C4 EEG only, which also
supply the occipital inputs, two EOG channels, chin EMG, ECG, oronasal airflow,
thoracic, abdominal and summed effort, SpO$_2$ and pulse, and its scored apneas
and hypopneas carry onset times. The EDF headers of this database carry no physical
unit, so EEG, EOG and EMG are scaled per recording to the median whole-night
amplitude of iSLEEPS, using no labels. UCD is used only for transfer.

\subsection{Signal Selection and Harmonization}

The iSLEEPS recordings contain 22 physiological and auxiliary traces. Fifteen
signals are retained for modeling and divided into neural and
cardiorespiratory groups. The neural group contains four EEG derivations
(C4:M1, C3:M2, O2:M1, and O1:M2), two EOG derivations (E1:M2 and E2:M2), and
the submental EMG. The cardiorespiratory group contains ECG, oronasal thermal
airflow, nasal pressure, thoracic effort, abdominal effort, summed respiratory
effort, SpO$_2$, and pulse.

Signals associated with periodic limb movement, snoring, plethysmography, body
position, ambient light, and device operation are not included in the model.
The snore channel is additionally excluded from respiratory prediction because
it is closely related to the airway information used during respiratory-event
assessment. Twelve iSLEEPS recordings lack at least one selected
cardiorespiratory channel. These patients are retained, and unavailable inputs
are zero-filled. The same
convention is used when a selected signal is unavailable in an external
dataset.

\subsection{Preprocessing and Epoch Labels}

All sleep-stage predictions are defined on 30-s epochs. For iSLEEPS and
ISRUC-Sleep, the respiratory annotations are mapped to the same epoch grid. An
epoch is assigned a positive respiratory label if a scored respiratory event begins
within it and a negative label otherwise. The primary
respiratory task is therefore prediction of respiratory-event onset within
a 30-s epoch, and the patient-level AHI is derived from these predictions
(Section~\ref{sec:ahimethod}).

iSLEEPS retains only the epochs scored as a sleep stage, so every cardiorespiratory
signal and every respiratory-event label is cut at the true recording time of each
retained epoch, keeping it on the same timeline as the EEG. The annotated AHI is the
number of apneas and hypopneas starting in scored sleep per hour of scored sleep.

The EEG, EOG, and EMG signals are resampled to $100$\,Hz for the engineered
descriptors of Section~\ref{sec:features}, and the four EEG and two EOG derivations
are resampled independently to $200$\,Hz for the LaBraM encoder, which does not
receive the EMG. The eight cardiorespiratory inputs are resampled to a common
$25$\,Hz grid by polyphase resampling and remain in their recorded physical units;
SpO$_2$, recorded at $4$\,Hz, is placed on the same grid without gaining temporal
resolution. Engineered features are standardized within each recording, which for
iSLEEPS is the patient, and the same per-recording procedure is applied to the
external datasets, from which no labels are used to refit parameters, select
thresholds, or adapt the reported iSLEEPS model.

\section{Proposed Method}
\label{sec:method}
\subsection{Problem Formulation}
A night is a sequence of $T$ epochs. For epoch $t$ we form a neural feature vector
$\mathbf{f}^{\mathrm{eeg}}_t\in\mathbb{R}^{388}$, formed by concatenating the $188$
engineered descriptors below with a $200$-dimensional embedding from LaBraM, a
pretrained EEG encoder~\cite{labram2024} adapted to sleep (Section~\ref{sec:encoder}), and a raw cardiorespiratory tensor
$\mathbf{x}^{\mathrm{car}}_t\in\mathbb{R}^{8\times750}$, the eight channels at $25$\,Hz.
The $14$ engineered cardiorespiratory descriptors defined below are used by the baseline
configuration against which the learned encoder is measured.

\begin{figure}[pos=H]
\centering
\includegraphics[width=\textwidth]{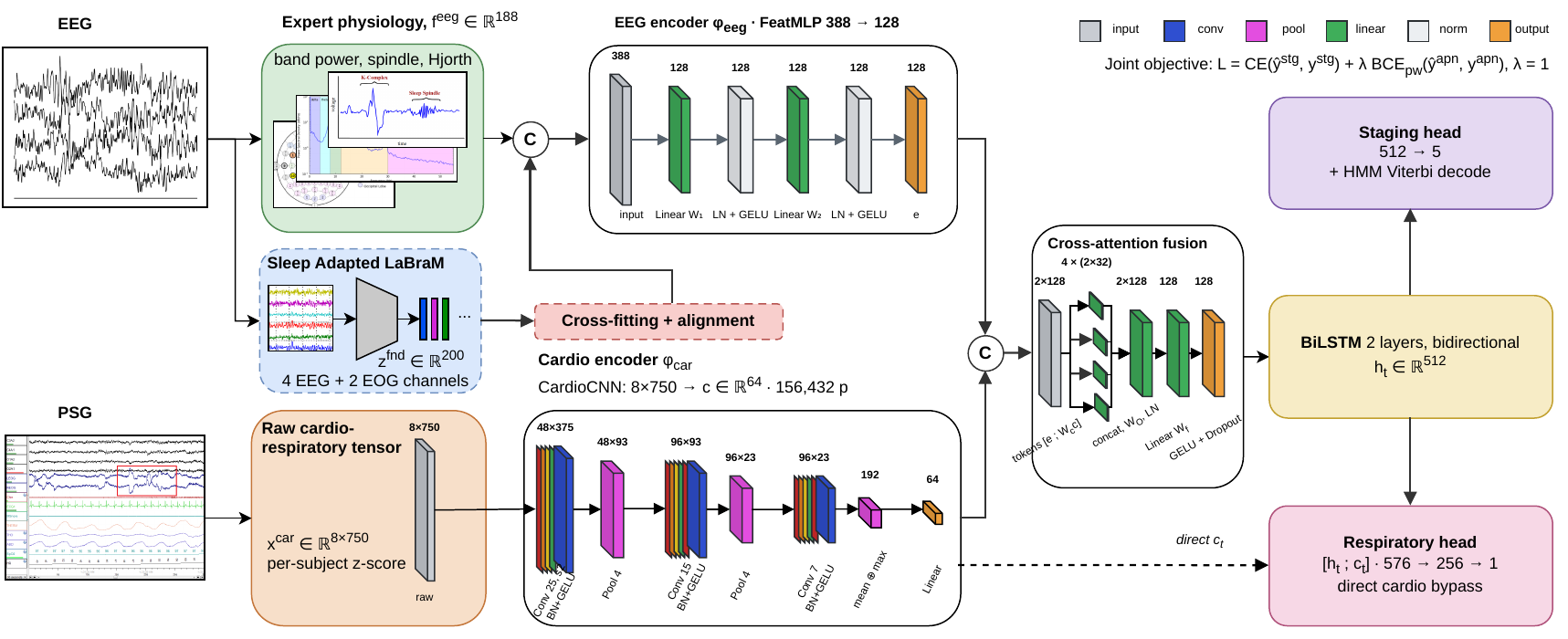}
\caption{The two-stream, multi-task model. The neural stream combines $188$ engineered
descriptors with the embedding of a LaBraM encoder adapted to sleep from EEG and EOG and
integrated by cross-fitting and alignment; the cardiorespiratory stream is a convolutional
encoder over eight raw channels. The streams are fused and read by a bidirectional LSTM over
a $20$-epoch window, which feeds a staging head and a respiratory head; the respiratory head
also reads the cardiorespiratory embedding directly, and the AHI is computed from the
respiratory scores over decoded sleep.}
\label{fig:arch}
\end{figure}

The encoder receives its input in the format used during pretraining. LaBraM fixes its input
length at $3000$ samples at $200$\,Hz; supplying the $100$\,Hz signal directly would compress time
by a factor of two, so that a $12$\,Hz spindle would appear at $24$\,Hz. Each $30$-s epoch is
therefore split into two $15$-s sub-windows that are encoded separately and mean-pooled. The four EEG derivations are presented
under the canonical names C4, C3, O2 and O1, the two EOG derivations as F7 and F8, the frontal
positions nearest the eyes, so that the encoder receives the ocular activity that separates REM
from Wake, and amplitudes are scaled to $\mu$V$/100$. We learn a subject-independent
map
\begin{equation}
f_\theta:\big(\mathbf{f}^{\mathrm{eeg}}_{1:T},\mathbf{x}^{\mathrm{car}}_{1:T}\big)\ \longmapsto\
\big(\hat y^{\mathrm{stg}}_{1:T},\ \hat y^{\mathrm{apn}}_{1:T}\big),
\end{equation}
where $\Delta^4=\{p\in\mathbb{R}^5:p_k\geq 0,\ \sum_k p_k=1\}$ is the four-simplex,
$\hat y^{\mathrm{stg}}_t\in\Delta^4$ is a distribution over the five stages and
$\hat y^{\mathrm{apn}}_t\in[0,1]$ scores whether epoch $t$ contains a respiratory
event (Figure~\ref{fig:arch}). We write $\hat y^{\mathrm{apn}}_t$ as a score and not as a
probability for the reason given in Section~\ref{sec:training}: the positive weighting in its
loss displaces the optimum away from the conditional event rate.

\subsection{Physiological Feature Representations}
\label{sec:features}
\paragraph{Neural features ($188$)}
Per channel we take absolute and relative band powers from the Welch spectrum $P(f)$ in delta
($0.5$--$4$), theta ($4$--$8$), alpha ($8$--$13$), sigma ($11$--$16$), and beta
($16$--$30$)\,Hz; the relative power in band $b$ is
\begin{equation}
\mathrm{RP}_b=\frac{\int_{f\in b}P(f)\,df}{\int_{0.5}^{30}P(f)\,df}.
\end{equation}
The conventional alpha and sigma bands overlap between $11$ and $13$\,Hz, so the five relative
powers are treated as correlated descriptors whose sum is unconstrained.
We add spectral entropy $H=-\sum_i p_i\log p_i$ with $p_i$ the normalized spectrum, the
$95\%$ spectral-edge frequency, the three Hjorth descriptors~\cite{hjorth1970eeg}, and
time-domain statistics. Sleep spindles are detected on the sigma-band analytic envelope,
the modulus of the analytic signal built from the Hilbert transform $\mathcal{H}$,
$e(t)=\lvert x_\sigma(t)+i\,\mathcal{H}\{x_\sigma\}(t)\rvert=\sqrt{x_\sigma(t)^2+\mathcal{H}\{x_\sigma\}(t)^2}$, with
density counted above a per-epoch threshold as a proxy for spindle density, and slow-wave
descriptors summarize slow-wave amplitude. Ocular-movement energy on the EOG and tonic EMG level
complete the set.

\paragraph{Cardiorespiratory features ($14$)}
From the seven cardiorespiratory channels other than nasal pressure, which is read only by the
learned encoder, we compute SpO$_2$ mean, minimum, variability, and
desaturation depth (median minus minimum); pulse mean and variability; ECG amplitude and
line-length; airflow amplitude and line-length; thoracic, abdominal, and summed effort
amplitudes; and the signed thoraco-abdominal synchrony
\begin{equation}
\mathrm{sync}_t=\mathrm{corr}(x^{\mathrm{tho}}_t,\ x^{\mathrm{abd}}_t),
\end{equation}
whose negative values mark the paradoxical motion of chest against abdomen in an obstructed
breath. These descriptors are read by the classical
respiratory baselines. Both feature sets are standardized per patient.

\paragraph{The learned cardiorespiratory encoder}
MM-Net does not read the $14$ descriptors. Its cardiorespiratory branch is a three-layer convolutional encoder over the raw $8\times750$ tensor, with wide kernels and aggressive downsampling because respiratory events are slow: at $25$\,Hz the first kernel of $25$ samples spans one second, and a hypopnea unfolds over $10$ to $30$\,s. Channel widths are
$48$, $96$ and $96$, each followed by batch normalization and GELU, with max-pooling by four
after the first two. Mean and max pooling over time are concatenated and projected by a
linear layer, LayerNorm, GELU and dropout to the same
$64$ dimensions as the cardiorespiratory feature MLP it replaces, so the encoder is a drop-in replacement and every
other part of the network is unchanged. The branch holds $156{,}432$ trainable parameters
with eight input channels ($155{,}232$ with seven), is trained end-to-end with the rest of the
network, and adds no external dependency.

\subsection{Sleep-Adapted EEG Encoder}
\label{sec:encoder}
\paragraph{Adaptation} Within each training fold, LaBraM-Base ($5.8$\,M parameters) is
fine-tuned with a linear staging head on single epochs of the fold's training patients, using
AdamW at $3\times10^{-5}$ for the encoder and $10^{-3}$ for the head, a one-cycle schedule
over three passes of the training epochs, batch size $64$ and mixed precision; the state with
the highest accuracy on the fold's validation patients is kept. Test patients never reach the
encoder.

\paragraph{Cross-fitting} An encoder fine-tuned on a patient embeds that patient in-sample. If the downstream model is trained on such embeddings and tested on out-of-sample ones, the two distributions differ, and the downstream model is fitted to representations of higher quality than those available at test time. Each fold's training patients are therefore split
into $K=3$ parts $P_1,P_2,P_3$, and an encoder $\phi_k$ is fine-tuned on the training patients
outside $P_k$; every training patient in $P_k$ is embedded by $\phi_k$, an encoder that never
saw it. The full-fold encoder $\phi$, fine-tuned on all training patients, embeds the
validation and test patients.

\paragraph{Alignment} Cross-fitting makes every embedding out of sample but places the
training patients in the spaces of $K$ different encoders, which do not coincide with the
space of $\phi$ used at test time. Each $\phi_k$ is mapped into the space of $\phi$ by a ridge
regression~\cite{hoerl1970ridge} fitted on the patients of $P_k$, for whom both embeddings
exist,
\begin{equation}
W_k=\arg\min_W\big\lVert(E^{\phi}_{P_k}-\mu^{\phi}_k)-(E^{\phi_k}_{P_k}-\mu^{\phi_k}_k)W\big\rVert_F^2
+\lambda n_k\lVert W\rVert_F^2,
\end{equation}
and the training embeddings become
$\tilde{\mathbf{e}}_t=(\mathbf{e}^{\phi_k}_t-\mu^{\phi_k}_k)W_k+\mu^{\phi}_k$, with $\lambda=0.01$
and $n_k$ the number of epochs in $P_k$. The map uses training patients only. Every training
embedding is thereby produced by an encoder that never saw its patient and expressed in the
coordinates the downstream model receives at test time; Section~\ref{sec:integration} shows that
the alignment is what allows the adapted encoder to benefit the downstream model.

\subsection{Two-Stream Encoders}
The neural stream is a two-layer perceptron with LayerNorm; the cardiorespiratory stream is the convolutional encoder described above, which emits a vector of the same width as the feature perceptron it replaced,
\begin{align}
e_t&=\mathrm{GELU}\!\big(\mathrm{LN}(W_2\,\mathrm{GELU}(\mathrm{LN}(W_1\mathbf{f}^{\mathrm{eeg}}_t))) \big)\in\mathbb{R}^{128},\\
c_t&=\mathrm{CardioCNN}\big(\mathbf{x}^{\mathrm{car}}_t\big)\in\mathbb{R}^{64}.
\end{align}
LayerNorm is used in place of BatchNorm because the recurrent windows are short and mix patients
within a batch.

\subsection{Fusion}
Two fusion operators were compared. Concatenation, used in the frozen-encoder configuration,
projects the joined embeddings back to the stream width,
\begin{equation}
z_t=\mathrm{GELU}\!\big(W_f\,[\,e_t\ ;\ c_t\,]\big)\in\mathbb{R}^{D},\qquad D=128.
\end{equation}
Cross-attention treats the two embeddings as tokens carrying learned modality-type
embeddings, refines them by four-head scaled dot-product attention so that each modality
reads the other, and projects the result to $z_t\in\mathbb{R}^{D}$. With uniform class weights
and the sleep-adapted encoder, cross-attention was selected on validation accuracy and is the
fusion of the final MM-Net (Table~\ref{tab:components}).

\subsection{Temporal Decoder and Two Heads}
Sleep is sequential and respiratory events cluster, so the fused vectors of an $L=20$ epoch
window ($10$ minutes) are read by a two-layer bidirectional LSTM. Training draws windows at a
stride of $L/2$ with masked zero-padding at the ends of each night; inference uses disjoint
windows, so every epoch is scored exactly once, and hidden states are not carried between
windows. Each direction runs the standard recurrence
\begin{align}
i_t&=\sigma(W_i[z_t,h_{t-1}]), & f_t&=\sigma(W_f[z_t,h_{t-1}]),\nonumber\\
o_t&=\sigma(W_o[z_t,h_{t-1}]), & g_t&=\tanh(W_g[z_t,h_{t-1}]),\\
s_t&=f_t\odot s_{t-1}+i_t\odot g_t, & h_t&=o_t\odot\tanh(s_t),\nonumber
\end{align}
and the forward and backward states of the $256$-unit layers are concatenated into
$h_t\in\mathbb{R}^{512}$. The
staging head is a linear map $\hat y^{\mathrm{stg}}_t=\mathrm{softmax}(W_s h_t)$. The
respiratory head reads $h_t$ together with a direct copy of the per-epoch cardiorespiratory
embedding $c_t$,
\begin{equation}
\hat y^{\mathrm{apn}}_t=\sigma\!\big(W_2\,\mathrm{GELU}(W_1[h_t;c_t])\big).
\end{equation}
MM-Net holds $2.85$ million trainable parameters. The sleep-adapted encoder that supplies
the $200$-dimensional neural embedding ($5.8$ million parameters) is fine-tuned in a separate
stage per fold (Section~\ref{sec:encoder}) and run once to cache the embeddings, so it is not
part of this count.

\subsection{Training and Decoding}
\label{sec:training}
The two tasks are trained together with equal weight, $\mathcal{L}=\mathcal{L}^{\mathrm{stg}}+\lambda\,\mathcal{L}^{\mathrm{apn}}$ with $\lambda=1$, combining a staging cross-entropy with class weights $w_k$ and a positive-weighted respiratory cross-entropy,
\begin{align}
\mathcal{L}^{\mathrm{stg}}&=-\sum_t\sum_{k}w_k\,y^{\mathrm{stg}}_{t,k}\log\hat y^{\mathrm{stg}}_{t,k},\\
\mathcal{L}^{\mathrm{apn}}&=-\sum_t\big[\pi\,a_t\log\hat y^{\mathrm{apn}}_t+(1-a_t)\log(1-\hat y^{\mathrm{apn}}_t)\big].
\end{align}
The final MM-Net uses uniform staging weights ($w_k=1$), selected on validation patients
together with the encoder (Table~\ref{tab:components}); the frozen-encoder configuration uses the
square root of inverse class frequency,
$w_k\propto\sqrt{N/(K n_k)}$ for $K=5$ classes with $n_k$ examples, which corrects the scarce
N1 and N3 stages. In both, the respiratory term is
positive-weighted by $\pi=n_-/n_+$ to match the $16\%$ event rate, which on this corpus is
$\pi\approx5.3$.

Against a true conditional rate $p$, the weighted binary term attains its minimum at
$q^{\star}=\pi p/[\pi p+(1-p)]>p$, so the respiratory output is treated as a score; the displacement is monotone, so AUC and average precision are
unaffected.

At inference the staging
posteriors are smoothed by a Viterbi pass whose transition matrix $A^{\mathrm H}$
and prior are counted from the training labels~\cite{chen2024hmm},
\begin{equation}
\delta_t(j)=\max_i\big[\delta_{t-1}(i)+\log A^{\mathrm H}_{ij}\big]+\log\hat y^{\mathrm{stg}}_t(j),
\end{equation}
recovered by back-tracking, which restores the natural persistence of sleep. Because the network
score, and not a generative likelihood $\log P(\mathbf{x}_t\mid s_t=j)$, occupies the emission
slot, the pass is a transition-regularized discriminative decoder that acts as a transition prior
on the output sequence; the respiratory head is left unsmoothed. Optimization
uses AdamW with cosine annealing and early stopping on a held-out set of patients.

\subsection{AHI Estimation}
\label{sec:ahimethod}
Because $\hat y^{\mathrm{apn}}_t$ scores event onsets, its sum over the epochs decoded as sleep
estimates the number of events, and dividing by the decoded sleep time gives a rate,
\begin{equation}
\widehat{\mathrm{AHI}}=k\cdot\frac{\sum_{t:\hat s_t\neq\mathrm{W}}\hat y^{\mathrm{apn}}_t}{\tfrac{30}{3600}\,\lvert\{t:\hat s_t\neq\mathrm{W}\}\rvert}.
\end{equation}
The positive weighting of the respiratory loss inflates the scores
(Section~\ref{sec:training}), so a single scale $k$ is fitted in each fold as the median ratio
of annotated to raw estimate over the fold's validation patients and applied unchanged to its
test patients. For transfer to another cohort, $k$ is the median of the ten iSLEEPS fold
values ($0.459$), and nothing is refitted on the external cohort. Severity follows the AASM
bands ($<5$, $5$--$15$, $15$--$30$ and $\geq30$ events/h).

\section{Experiments}
\label{sec:results}
\subsection{Protocol and Metrics}
All results use ten-fold, patient-independent cross-validation: no patient appears in both
training and test, and within each training split a disjoint set of patients is held out for
early stopping and model selection. Staging is scored by accuracy, macro-F1
$\tfrac{1}{K}\sum_k\mathrm{F1}_k$, and Cohen's $\kappa$. The respiratory task is scored by the
area under the ROC curve (AUC) and average precision (AP), which suits the $16\%$ event rate,
and the AHI estimate by Spearman's $\rho$, the mean absolute error (MAE), severity-band accuracy
and linearly weighted $\kappa$.

MM-Net and its encoder configurations are trained with three seeds ($42$, $1$, $7$) on the same
folds. The single-model figure is the mean over the thirty fold-values, ten folds by three seeds; the
three-seed figure averages the softmax outputs of the three seeds before decoding. Every number
is a mean over the ten folds with its standard deviation across folds. Paired comparisons use
the two-sided Wilcoxon signed-rank test on the ten fold means, because the three seeds within a
fold train and test on the same patients; with ten folds the smallest attainable $p$ is $0.002$,
and Holm correction is applied within each family of comparisons.

The frozen-encoder configuration, the initial configuration of the same architecture, serves as
the reference for the encoder analyses. It shares the streams, decoder and heads of MM-Net but
reads a frozen LaBraM embedding of the EEG, seven cardiorespiratory channels, concatenation
fusion and square-root class weights. Its stream, decoder and regularization settings, which MM-Net
inherits, were chosen among forty-nine candidates; repeating that selection inside a nested
procedure, with five outer patient-independent folds whose test patients take no part in the
choice, gives $0.734 \pm 0.009$ accuracy against $0.739$, a selection optimism of $0.006$, about a
third of the fold-to-fold standard deviation. The encoder, fusion operator and class weighting of
the final MM-Net were selected on validation patients only.

\subsection{Implementation and Hyperparameters}
The model is implemented in PyTorch and trained on NVIDIA RTX~2060 (6\,GB) and RTX~3060\,Ti
(8\,GB) GPUs. Table~\ref{tab:hyper} lists the training configuration, which is identical across
folds and cohorts. The reimplemented baselines are trained with the optimizer, batch size and
schedule of their own papers, because imposing our settings could disadvantage architectures
tuned for other settings.

\begin{table}[H]
\caption{Training configuration of the final MM-Net, shared across all folds and cohorts. The
frozen-encoder configuration uses the same settings with a frozen encoder, concatenation and square-root
class weights. Each reimplemented baseline is trained at its own published settings.}
\label{tab:hyper}
\centering
\small
\adjustbox{max width=\textwidth}{%
\begin{tabular}{ll}
\toprule
Setting & Value \\
\midrule
Framework / hardware & PyTorch, NVIDIA RTX~2060 / 3060\,Ti \\
\multicolumn{2}{@{}l}{\emph{Sleep-adapted encoder (per fold)}} \\
Encoder & LaBraM-Base, $4$ EEG $+$ $2$ EOG \\
Encoder LR (head) & $3\times10^{-5}$ ($10^{-3}$), one-cycle \\
Passes, batch, weight decay & $3$, $64$, $0.05$ \\
Cross-fitting, alignment & $K=3$, ridge $\lambda=0.01$ \\
\multicolumn{2}{@{}l}{\emph{MM-Net}} \\
Optimizer & AdamW \\
Learning rate & $3\times10^{-4}$ \\
Weight decay & $1\times10^{-4}$ \\
LR schedule & cosine annealing \\
Sequence length $L$ & $20$ epochs ($10$\,min) \\
Batch size & $32$ \\
Max epochs / patience & $45$ / $8$ \\
Gradient clipping & $5.0$ \\
Fusion & cross-attention, $4$ heads \\
Staging class weights & uniform \\
Respiratory pos.\ weight & $\pi=n_-/n_+$ \\
Loss balance $\lambda$ & $1.0$ \\
Staging post-processing & HMM Viterbi smoothing \\
Cross-validation & $10$-fold, patient-independent \\
Random seeds & $42$, $1$, $7$ \\
\bottomrule
\end{tabular}}
\end{table}

\subsection{Compared Methods}
We compare the model against deep networks evaluated on the same folds: convolutional and
recurrent networks trained from scratch (DeepSleepNet, AttnSleep, a four-channel CNN),
six further published architectures we reimplemented for this comparison (U-Time, a
fully convolutional encoder--decoder; TinySleepNet, a compact convolutional--recurrent model;
SleepTransformer, a purely attentional one; and 1D-ResNet-SE-LSTM, Micro SleepNet and
AT-BiLSTM), the released SE-ResNet-18 implementation from~\cite{bose2026gap}, a
StagingSeqNet pretrained on Sleep-EDF and fine-tuned on iSLEEPS, and a raw-signal convolutional network on all $14$ channels,
alongside the published single-channel baselines for the corpus~\cite{maiti2026isleeps}. For the
respiratory task, a desaturation rule, logistic regression and gradient boosting read the $14$
engineered cardiorespiratory descriptors, and two raw-signal detectors read the waveforms
(Section~\ref{sec:noflow}). On ISRUC-I, the reimplemented staging baselines are retrained on
MM-Net's folds (Section~\ref{sec:external}). All numbers are
patient-independent. All use ten folds except the Sleep-EDF-pretrained StagingSeqNet, which uses five; hidden-Markov decoding is applied where the corresponding model provides it.

\begin{table}[H]
\caption{Patient-independent sleep-staging comparison on iSLEEPS. Values are mean $\pm$ SD where fold-level results are available. $^{\dagger}$ denotes reimplemented models, $^{\ddagger}$ a literature-reported result under its original protocol, and $^{\ast}$ models ported from released code. Bold indicates the best result among models evaluated under our protocol.}
\label{tab:bench}
\centering
\adjustbox{max width=\textwidth}{%
\begin{tabular}{@{}l l c c c c c@{}}
\toprule
 & & & Healthy & \multicolumn{3}{c}{iSLEEPS (stroke)} \\
\cmidrule(lr){4-4}\cmidrule(lr){5-7}
Model & Input & Params\,$\downarrow$ & Acc & Acc\,$\uparrow$ & Macro-F1\,$\uparrow$ & $\kappa$\,$\uparrow$ \\
\midrule
\multicolumn{7}{@{}l}{\emph{Comparative models evaluated on iSLEEPS}}\\[1pt]
LSTM, our reimplementation$^{\dagger}$ & 1\,EEG\,$+$\,EOG & $1.16$\,M & $0.820 \pm 0.062$ & $0.720 \pm 0.032$ & $0.626 \pm 0.028$ & $0.599 \pm 0.043$ \\
CNN-ResNet18~\cite{bose2026gap}$^{\ddagger}$ & 1\,EEG & $\sim$11\,M & --- & 0.617 & 0.544 & 0.480 \\
SleepStageNet~\cite{alhazmi2026sleepstagenet}$^{\ddagger}$ & 2\,EEG\,$+$\,2\,EOG\,$+$\,EMG & $1.08$\,M & --- & $0.739 \pm 0.022$ & $0.673 \pm 0.020$ & $0.634 \pm 0.029$ \\
SE-ResNet-18\,+\,BiLSTM~\cite{bose2026gap}$^{\ast}$ & 1\,EEG & $11.32$\,M & $0.829 \pm 0.056$ & $0.696 \pm 0.038$ & $0.606 \pm 0.032$ & $0.572 \pm 0.046$ \\
DeepSleepNet~\cite{supratak2017deepsleepnet}  & 4\,EEG  & $3.72$\,M  & $0.823 \pm 0.050$ & $0.669 \pm 0.031$ & $0.624 \pm 0.026$ & $0.558 \pm 0.035$ \\
U-Time~\cite{perslev2019utime}$^{\dagger}$ & 1\,EEG & $1.07$\,M & $0.791 \pm 0.062$ & $0.646 \pm 0.064$ & $0.524 \pm 0.057$ & $0.506 \pm 0.082$ \\
AttnSleep~\cite{eldele2021attnsleep}          & 1\,EEG  & $\sim$0.5\,M & $0.833 \pm 0.042$ & $0.716 \pm 0.042$ & $0.600 \pm 0.043$ & $0.593 \pm 0.058$ \\
TinySleepNet~\cite{supratak2020tinysleepnet}$^{\dagger}$ & 1\,EEG & $1.45$\,M & $0.824 \pm 0.045$ & $0.676 \pm 0.050$ & $0.587 \pm 0.034$ & $0.548 \pm 0.062$ \\
SleepTransformer~\cite{sleeptransformer2022}$^{\dagger}$ & 1\,EEG & $2.79$\,M & $0.823 \pm 0.033$ & $0.683 \pm 0.026$ & $0.602 \pm 0.030$ & $0.555 \pm 0.032$ \\
1D-ResNet-SE-LSTM~\cite{li2023resnetselstm}$^{\dagger}$ & 1\,EEG & $0.65$\,M & $0.832 \pm 0.036$ & $0.708 \pm 0.038$ & $0.631 \pm 0.033$ & $0.592 \pm 0.050$ \\
Micro SleepNet~\cite{liu2023microsleepnet}$^{\dagger}$ & 1\,EEG & $0.03$\,M & $0.800 \pm 0.041$ & $0.647 \pm 0.049$ & $0.558 \pm 0.028$ & $0.510 \pm 0.055$ \\
AT-BiLSTM~\cite{fu2021atbilstm}$^{\dagger}$ & 1\,EEG & $0.92$\,M & $0.835 \pm 0.046$ & $0.692 \pm 0.040$ & $0.612 \pm 0.030$ & $0.570 \pm 0.051$ \\
CNN                                           & 4\,EEG  & $0.42$\,M & $0.806 \pm 0.041$ & $0.670 \pm 0.052$ & $0.618 \pm 0.044$ & $0.553 \pm 0.062$ \\
StagingSeqNet, Sleep-EDF pretrained           & 2\,EEG\,+\,EOG\,+\,EMG & $1.08$\,M & --- & $0.623 \pm 0.020$ & 0.582 & 0.488 \\
Raw multimodal CNN                            & 14\,ch  & 0.95\,M      & --- & $0.677 \pm 0.035$ & $0.633 \pm 0.025$ & $0.563 \pm 0.041$ \\
Huttunen~et~al.~\cite{huttunen2023simultaneous}$^{\ast}$ & 4\,ch & $3.80$\,M & --- & $0.599 \pm 0.060$ & $0.495 \pm 0.067$ & $0.440 \pm 0.078$ \\
\addlinespace[2pt]\midrule
\multicolumn{7}{@{}l}{\emph{Sleep-adapted encoder alone (one epoch, no temporal model)}}\\[1pt]
LaBraM, frozen $+$ logistic regression & 4\,EEG & $5.8$\,M & --- & $0.647$ & --- & --- \\
LaBraM, sleep-adapted                  & 4\,EEG\,$+$\,2\,EOG & $5.8$\,M & --- & $0.732$ & --- & --- \\
\addlinespace[2pt]\midrule
\multicolumn{7}{@{}l}{\emph{Proposed multi-task model}}\\[1pt]
MM-Net, single model         & 7\,neural $+$ 8\,cardio & 2.85\,M & $0.864 \pm 0.030$ & $0.756 \pm 0.013$ & $0.661 \pm 0.019$ & $0.647 \pm 0.015$ \\
\textbf{MM-Net, three-seed}  & 7\,neural $+$ 8\,cardio & 2.85\,M & $\mathbf{0.864 \pm 0.030}$ & $\mathbf{0.763 \pm 0.014}$ & $\mathbf{0.664 \pm 0.020}$ & $\mathbf{0.656 \pm 0.016}$ \\
\bottomrule
\end{tabular}}
\end{table}

\subsection{Staging Benchmark}
Table~\ref{tab:bench} compares MM-Net with deep staging architectures retrained on the same
folds. These reach $0.60$--$0.72$ staging accuracy, whereas MM-Net reaches $\mmAccS$ as a single
model and $\mmAcc$ as a three-seed average, with macro-F1 $\mmMf$ and $\kappa$ $\mmK$. As a
single model, MM-Net is higher in accuracy by $+0.036$ (corpus LSTM) to $+0.157$
(Huttunen~\emph{et al.}) and is higher in nine or ten of the ten folds for every comparison; all
thirteen differences remain significant after Holm correction (Holm $p\leq0.025$). The three-seed
MM-Net is ahead of each of the eight reimplemented and ported architectures in all ten folds
(Holm $p=0.016$, the minimum attainable for eight comparisons). Differences in macro-F1 and
$\kappa$ are larger than those in accuracy, consistent with their reduced dependence on the
majority classes. The sleep-adapted encoder used on its own stages single epochs at $0.732$,
already above every retrained architecture; within MM-Net, together with the engineered
descriptors, the cardiorespiratory stream and the temporal decoder, it yields $\mmAccS$, a further
$0.024$ in eight folds of ten ($p=0.014$).

The closest task-matched baseline is the joint model of Huttunen~\emph{et al.}
\cite{huttunen2023simultaneous}, ported from the authors' released implementation and evaluated
on the same folds. It stages at $0.599$, and MM-Net is ahead in all ten folds (Wilcoxon
$p=0.002$). On ISRUC-I, where both models are
trained and evaluated on the same respiratory labels, MM-Net is also ahead on respiratory
detection (Section~\ref{sec:external}).

SleepStageNet~\cite{alhazmi2026sleepstagenet} is quoted under its own protocol, $95$ iSLEEPS
recordings with five channels and ten subject-independent folds. The CNN-ResNet18 result
of~\cite{bose2026gap} is quoted under that study's $N=100$ protocol,
which includes the duplicate recording excluded here; the same study's SE-ResNet-18\,+\,BiLSTM,
retrained from the released implementation on our folds, reaches $0.696$ against the published
$0.747$, and Table~\ref{tab:bench} uses the same-fold rerun. The corpus LSTM appears as our
reimplementation on the same folds. The Healthy column gives each
model's subject-independent accuracy on Sleep-EDF, where MM-Net reaches $0.864$ (macro-F1
$0.791$, $\kappa$ $0.810$), the highest healthy-sleep accuracy in the table; Sleep-EDF carries
none of the cardiorespiratory channels the model reads, so this figure exercises the neural
stream alone. Dispersion is reported wherever it is defined: across five folds for the
Sleep-EDF-pretrained StagingSeqNet, across the ten fold values of a single seed for the retrained
and reimplemented architectures, and across the thirty fold values, ten folds by three seeds, for
AttnSleep and MM-Net.

\begin{figure}[pos=H]
\centering
\includegraphics[width=0.78\textwidth]{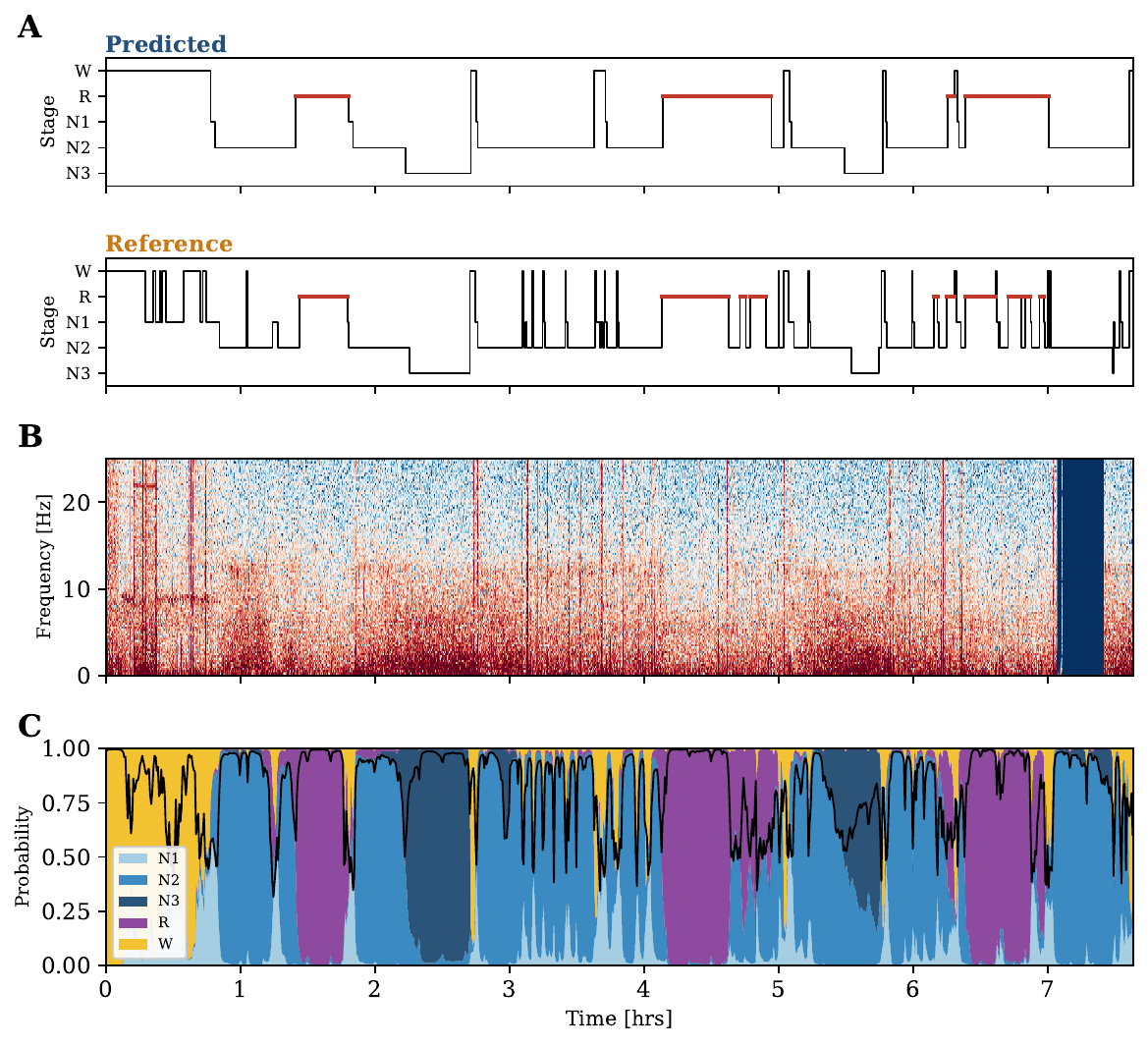}
\caption{An illustrative held-out night (patient SN90, three-seed MM-Net, staging accuracy
$0.83$). \textbf{(A)} Predicted and reference hypnograms, REM highlighted. \textbf{(B)} EEG
spectrogram, C4 derivation, $0$--$25$\,Hz. \textbf{(C)} Per-epoch stage-probability
ribbon; the black line is the maximum posterior.}
\label{fig:hypno}
\end{figure}

\begin{figure}[pos=H]
\centering
\includegraphics[width=0.42\textwidth]{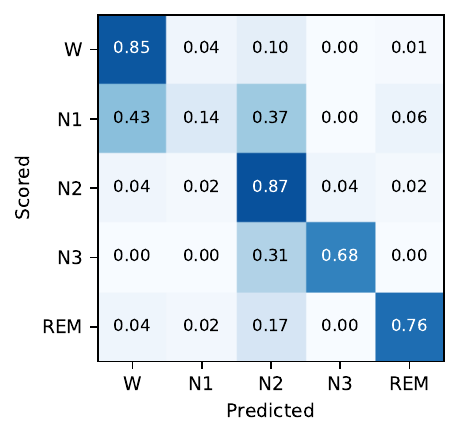}
\caption{Row-normalized confusion matrix for MM-Net (three-seed, pooled ten-fold test epochs,
hidden-Markov operating point). Errors concentrate on N1, which is absorbed into Wake and N2.}
\label{fig:confusion}
\end{figure}

Errors concentrate on N1 (Table~\ref{tab:perclass}, Figure~\ref{fig:confusion}). MM-Net recovers Wake, N2 and REM at F1
$0.79$--$0.82$ and N3 at $0.72$, whereas N1, the transient stage between wake and sleep, reaches
$0.20$ and is absorbed mainly into Wake and N2; part of N3 is read as N2. N1 is the most difficult
stage on this cohort for every evaluated model. A representative held-out night is shown in Figure~\ref{fig:hypno}.

\begin{table}[H]
\caption{Per-class staging F1 (Wake, N1, N2, N3, REM), ten-fold patient-independent, at the
hidden-Markov operating point (MM-Net: three-seed, pooled test epochs).}
\label{tab:perclass}
\centering
\setlength{\tabcolsep}{6pt}
\begin{tabular}{lccccc}
\toprule
Model & W\,$\uparrow$ & N1\,$\uparrow$ & N2\,$\uparrow$ & N3\,$\uparrow$ & R\,$\uparrow$ \\
\midrule
MM-Net    & 0.81 & 0.20 & 0.82 & 0.72 & 0.79 \\
\bottomrule
\end{tabular}
\end{table}

\subsection{Integrating the Sleep-Adapted Encoder}
\label{sec:integration}
\begin{figure}[pos=H]
\centering
\includegraphics[width=0.78\textwidth]{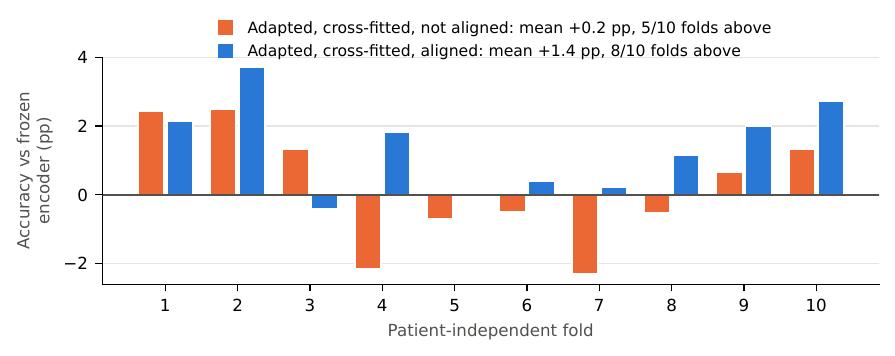}
\caption{Per-fold staging accuracy of MM-Net with the sleep-adapted encoder, relative to the
frozen-encoder reference (three-seed, ten patient-independent folds). The training embeddings are
cross-fitted in both cases; aligning them to the deployed encoder turns an inconsistent effect into
a gain in eight folds of ten.}
\label{fig:alignfolds}
\end{figure}

Table~\ref{tab:components} builds MM-Net from the frozen-encoder configuration one change at a
time, every row on the same folds and seeds. Uniform class weights and cross-attention fusion
contribute $+0.009$ together for the single model. The largest change comes from the encoder.
Adapting LaBraM to sleep raises its own single-epoch accuracy from $0.647$ to $0.709$, in all ten
folds, yet a sleep-adapted encoder whose training embeddings are cross-fitted but not aligned
moves the three-seed accuracy of the downstream model by only $+0.002$ ($p=0.63$) and falls below
the frozen encoder in five folds of ten, by up to $0.023$. Aligning the embedding spaces
turns the same encoder into a gain of $+0.014$ ($0.755\rightarrow0.769$, $p=0.020$), and of
$+0.015$ for the single model in nine folds of ten ($p=0.027$). Figure~\ref{fig:alignfolds} shows
the effect fold by fold: without alignment the adapted encoder moves accuracy in either direction,
whereas with alignment it raises accuracy in eight folds of ten, by up to $3.7$ percentage
points. The downstream benefit of adaptation therefore appears once the training embeddings share
the coordinates of the encoder used at test time.

\begin{figure}[pos=H]
\centering
\includegraphics[width=0.66\textwidth]{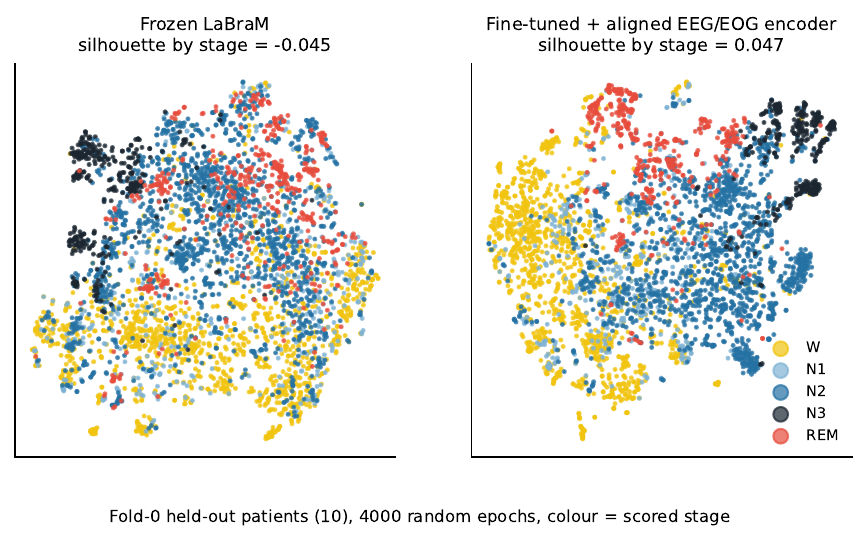}
\caption{t-SNE~\cite{vandermaaten2008tsne} of $4000$ epochs from the ten held-out patients of
one fold, colored by scored stage. Left: frozen LaBraM. Right: the sleep-adapted, aligned
EEG/EOG encoder used in MM-Net.}
\label{fig:tsne}
\end{figure}

The complete MM-Net adds three elements that serve REM and the respiratory head more than overall
staging: the ocular channels in the encoder, which raise the encoder's own accuracy from $0.709$
to $0.732$, nasal pressure, and the epoch alignment of Section~\ref{sec:data}. With all three,
staging is $\mmAcc$, and MM-Net uses the EEG and EOG encoder in every cohort of this study. The same integration carries over to an independent cohort: on ISRUC-I the complete model
is ahead of the frozen-encoder configuration in nine folds of ten ($+0.018$, $p=0.004$;
Section~\ref{sec:external}). Figure~\ref{fig:tsne} shows what adaptation changes in the
representation: held-out epochs that the frozen encoder spreads across one mixed cloud form
stage-specific clusters after adaptation, with Wake, REM and N3 separated (silhouette by stage
$-0.045\rightarrow0.047$).

\begin{table}[H]
\caption{From the frozen-encoder configuration to MM-Net on iSLEEPS, one change per row (upper
block), and the complete model (lower block); same ten folds and three seeds throughout. $p$:
Wilcoxon on fold means against the row marked ref.}
\label{tab:components}
\centering
\small
\setlength{\tabcolsep}{3pt}
\adjustbox{max width=\textwidth}{%
\begin{tabular}{@{}lccc@{}}
\toprule
Configuration & Single & 3-seed & $p$ \\
\midrule
Frozen encoder, class weights, concatenation & 0.738 & 0.749 & \\
\ \ $+$ uniform class weights & 0.744 & 0.754 & \\
\ \ $+$ cross-attention fusion & 0.747 & 0.755 & ref. \\
\ \ $+$ sleep-adapted encoder, cross-fitted, not aligned & 0.752 & 0.757 & 0.63 \\
\ \ $+$ ridge alignment (7 channels) & \textbf{0.762} & \textbf{0.769} & \textbf{0.020} \\
\midrule
MM-Net: EEG$+$EOG encoder, 8 channels, epoch-aligned & 0.756 & 0.763 & \\
\bottomrule
\end{tabular}}
\end{table}

\subsection{Respiratory-Event Detection}
\label{sec:noflow}
\begin{figure}[pos=H]
\centering
\includegraphics[width=0.78\textwidth]{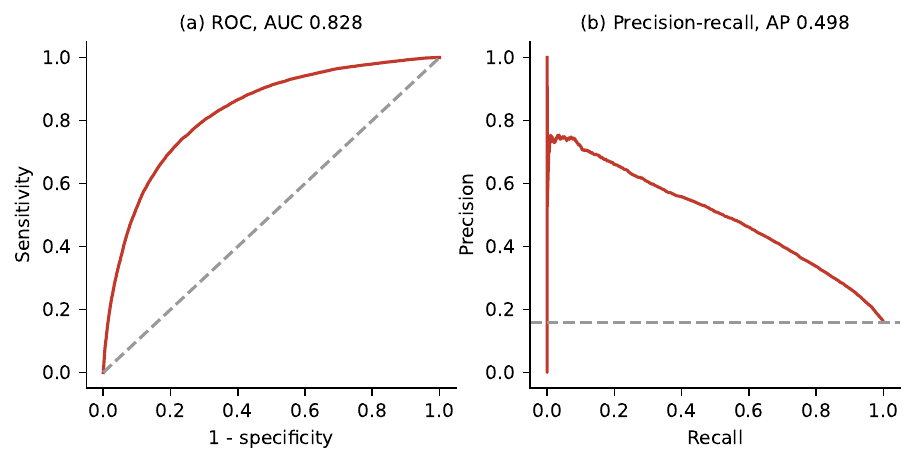}
\caption{Respiratory-event detection by MM-Net (three-seed), pooled over the ten
patient-independent test folds. \textbf{(a)} ROC curve. \textbf{(b)} Precision--recall; the
dashed line is the $0.160$ event prevalence, the no-skill floor.}
\label{fig:rocpr}
\end{figure}

Table~\ref{tab:resp} compares the respiratory output with established detectors on the same folds
and epoch labels. A clinical desaturation rule reaches $0.595$ AUC, and logistic regression and
gradient boosting on the $14$ engineered cardiorespiratory descriptors reach $0.603$ and $0.673$;
a random classifier scores $0.160$ AP, the event prevalence. Two raw-signal detectors, each a
per-epoch convolutional encoder over the $25$\,Hz waveform followed by a bidirectional LSTM across
$20$-epoch windows and trained with the same labels, folds, early-stopping rule and positive-class
weighting, read summed respiratory effort alone, matching the input of
Nassi~\emph{et al.}~\cite{nassi2022effortbelt} ($0.602$ AUC), or all seven raw cardiorespiratory
channels ($0.733$; one seed each). MM-Net reaches $\mmAuc$ AUC and $\mmAp$ AP, $0.152$ AUC and
$0.197$ AP above gradient boosting. Detection is strongest on the most severe events: $0.810$ AUC
on hypopneas, which form $82\%$ of event epochs, $0.898$ on obstructive and $0.946$ on central
apneas. Read at fixed sensitivity on the pooled held-out predictions (Figure~\ref{fig:rocpr}),
MM-Net holds $0.70$ specificity at $0.80$ sensitivity and $0.53$ at $0.90$.

\begin{table}[H]
\caption{Respiratory-event detection on iSLEEPS. (A) Comparison with classical and raw-signal baselines on the same folds ($^{\S}$single seed) and (B) MM-Net AUC by respiratory-event type. Bold denotes the best value in each comparison.}
\label{tab:resp}
\centering
\small
\begin{tabular}{lcc}
\toprule
\multicolumn{3}{l}{\textit{(a) Detection method}}\\
\midrule
Method & AUC\,$\uparrow$ & AP\,$\uparrow$ \\
\midrule
Desaturation rule    & 0.595 & 0.207 \\
Logistic regression  & 0.603 & 0.211 \\
Gradient boosting    & 0.673 & 0.290 \\
Raw effort belt$^{\S}$~\cite{nassi2022effortbelt} & 0.602 & 0.252 \\
Raw seven channels$^{\S}$ & 0.733 & 0.366 \\
MM-Net, single model & 0.815 & 0.471 \\
\textbf{MM-Net, three-seed} & \textbf{0.825} & \textbf{0.487} \\
\bottomrule
\end{tabular}

\vspace{4pt}
\begin{tabular}{lcc}
\toprule
\multicolumn{3}{l}{\textit{(b) MM-Net AUC by event type}}\\
\midrule
Event type & Positive epochs & AUC\,$\uparrow$ \\
\midrule
Hypopnea    & 11{,}921 & 0.810 \\
Obstructive & 1{,}722  & 0.898 \\
Central     & 832      & \textbf{0.946} \\
\bottomrule
\end{tabular}
\end{table}

\subsection{Joint Training and Stream Contributions}
\label{sec:joint}
The joint objective is compared with matched single-task models that share the architecture,
folds, seeds, optimizer, training budget and decoding of the joint model and differ only in the
surviving term of $\mathcal{L}=\mathcal{L}^{\mathrm{stg}}+\lambda\mathcal{L}^{\mathrm{apn}}$
(Table~\ref{tab:joint}). Joint training improves staging where disordered breathing disrupts
sleep. In epochs containing a respiratory-event onset, MM-Net stages at $0.721$ against $0.705$
for the staging-only model, in nine folds of ten ($p=0.014$), and the gain persists within two
epochs of an event ($0.710$ against $0.701$, $p=0.020$). In patients with severe sleep apnea
(AHI $\geq30$, $n=38$), accuracy rises from $0.758$ to $0.768$ ($p=0.020$), and the stages
that respiratory events fragment improve: N3 F1 from $0.695$ to $0.707$ ($p=0.049$) and N2 F1
from $0.811$ to $0.816$ ($p=0.037$). Across the cohort, per-patient accuracy rises from $0.756$
to $0.762$ ($60$ of $99$ patients, $p=0.013$). Joint training also improves respiratory-event
detection: in the frozen-encoder configuration, respiratory AUC rises from $0.784$ to $0.796$ in
nine folds of ten ($p=0.006$), and in MM-Net the respiratory AUC and AP are $0.815$ and $0.471$
against $0.814$ and $0.466$ for the respiratory-only model. Summing the joint model's respiratory
scores over its own decoded sleep gives an AHI estimate closer to the annotated index than
combining the two single-task models (Spearman $\rho$ $0.765$ against $0.734$; mean absolute
error $10.4$ against $11.1$ events/h).

\begin{table}[H]
\caption{Joint multi-task training against matched single-task models with the same architecture,
folds, seeds, training budget and decoding. MM-Net staging: three-seed ensemble; respiratory AUC
and AP: mean over seeds. $p$: Wilcoxon on the ten fold means, or on patients for the severe-apnea
group and the per-patient accuracy. The single-task AHI combines the stages of the staging-only
model with the scores of the respiratory-only model.}
\label{tab:joint}
\centering
\small
\setlength{\tabcolsep}{5pt}
\adjustbox{max width=\textwidth}{%
\begin{tabular}{@{}llccc@{}}
\toprule
 & Measure & Single-task & Joint & $p$ \\
\midrule
\multirow{6}{*}{Staging} & Accuracy, epochs with an event onset & $0.705$ & $\mathbf{0.721}$ & $0.014$ \\
 & Accuracy, within two epochs of an event & $0.701$ & $\mathbf{0.710}$ & $0.020$ \\
 & Accuracy, severe sleep apnea (AHI $\geq30$) & $0.758$ & $\mathbf{0.768}$ & $0.020$ \\
 & N3 F1 & $0.695$ & $\mathbf{0.707}$ & $0.049$ \\
 & N2 F1 & $0.811$ & $\mathbf{0.816}$ & $0.037$ \\
 & Accuracy, per patient & $0.756$ & $\mathbf{0.762}$ & $0.013$ \\
\midrule
\multirow{3}{*}{Respiratory} & AUC, frozen-encoder configuration & $0.784$ & $\mathbf{0.796}$ & $0.006$ \\
 & AUC & $0.814$ & $\mathbf{0.815}$ & --- \\
 & AP & $0.466$ & $\mathbf{0.471}$ & --- \\
\midrule
\multirow{2}{*}{AHI} & Spearman $\rho$ & $0.734$ & $\mathbf{0.765}$ & --- \\
 & Mean absolute error (events/h) & $11.1$ & $\mathbf{10.4}$ & --- \\
\bottomrule
\end{tabular}}
\end{table}

Signal interventions in the frozen-encoder configuration show that each stream carries the output
its physiology is closest to (Table~\ref{tab:ablate}). Removing the EEG and retraining lowers
staging accuracy by $0.050$, and removing the whole cardiorespiratory stream lowers respiratory
AUC by $0.118$, the largest effects of the modality-ablation grid. Within the cardiorespiratory
stream, respiratory effort is the channel the detector relies on most ($-0.028$ AUC without it),
consistent with the paradoxical thoraco-abdominal motion that marks an obstructive event as it
happens. Removing airflow, the signal from which events are scored, leaves detection essentially
unchanged ($-0.004$), so the respiratory output is supported by the physiological correlates of
disordered breathing and not by the scoring channel alone.

\begin{table}[H]
\caption{Modality removal and retraining (frozen-encoder configuration; ten folds, three seeds).
Each row removes one modality and reports the change in the output that modality serves, relative
to the full configuration (staging accuracy $0.736$, respiratory AUC $0.796$). Holm $p$: Wilcoxon
on the ten fold means, Holm-corrected across the nine ablations of the grid.}
\label{tab:ablate}
\centering
\small
\setlength{\tabcolsep}{6pt}
\begin{tabular}{llcc}
\toprule
Modality removed & Output & Change & Holm $p$ \\
\midrule
EEG (neural)            & Staging accuracy & $-0.050$ & $0.018$ \\
All cardiorespiratory   & Respiratory AUC  & $-0.118$ & $0.018$ \\
Respiratory effort      & Respiratory AUC  & $-0.028$ & $0.031$ \\
Airflow                 & Respiratory AUC  & $-0.004$ & $0.65$ \\
\bottomrule
\end{tabular}
\end{table}

\subsection{AHI Estimation}
\label{sec:ahi}
\begin{figure}[pos=H]
\centering
\includegraphics[width=0.66\textwidth]{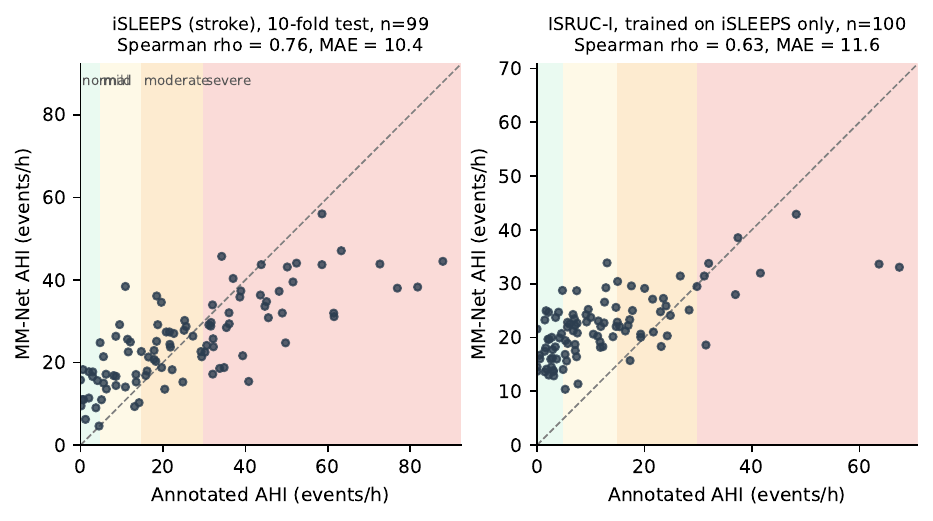}
\caption{MM-Net AHI estimate against the annotated AHI. Left: iSLEEPS, ten-fold test patients.
Right: ISRUC-I, MM-Net trained on iSLEEPS only and applied without retraining
(Section~\ref{sec:external}). Shading marks the AASM severity bands; the dashed line is
identity.}
\label{fig:ahi}
\end{figure}

Summed over decoded sleep, the respiratory scores estimate the AHI
(Section~\ref{sec:ahimethod}) with Spearman $\rho=\mmRho$ against the annotated index over the
$99$ test patients, a mean absolute error of $10.4$ events/h, and $54\%$ of patients in the
correct AASM severity band (weighted $\kappa$ $0.45$; Table~\ref{tab:ahi}, Figure~\ref{fig:ahi}). The clinical AHI
recorded with the corpus agrees with the annotated index at $\rho=0.687$, and against that
clinical index MM-Net reaches $\rho=0.633$. In the expected-count formulation, summing the onset
scores over decoded sleep counts closely spaced events separately, whereas thresholding the scores
into runs would merge them, and normalizing by decoded sleep time matches the denominator of the
clinical index. Staging remains stable across sleep-disordered-breathing severity: accuracy is $0.784$ in patients
with a normal clinical AHI and $0.754$, $0.764$ and $0.758$ in the mild, moderate and severe
groups, with no significant ordered trend (Jonckheere--Terpstra $p=0.20$;
Table~\ref{tab:severity}), so the model stages the most severely affected patients nearly as well
as those without sleep-disordered breathing.

\begin{table}[H]
\caption{AHI estimation on iSLEEPS ($99$ test patients, stages decoded by MM-Net). Severity:
AASM bands; $\kappa_w$ linearly weighted. Last column: Spearman $\rho$ against the clinical AHI
recorded with the corpus. The last row gives the agreement of the clinical AHI with the
annotated AHI.}
\label{tab:ahi}
\centering
\small
\setlength{\tabcolsep}{3pt}
\adjustbox{max width=\textwidth}{%
\begin{tabular}{@{}lccccc@{}}
\toprule
Estimator & $\rho$\,$\uparrow$ & MAE\,$\downarrow$ & Sev.\ acc\,$\uparrow$ & $\kappa_w$\,$\uparrow$ & $\rho_{\mathrm{clin}}$\,$\uparrow$ \\
\midrule
\textbf{MM-Net} & \textbf{0.765} & \textbf{10.4} & \textbf{0.54} & \textbf{0.45} & \textbf{0.633} \\
\midrule
Clinical vs.\ annotated AHI & 0.687 & & & & \\
\bottomrule
\end{tabular}}
\end{table}

\begin{table}[H]
\caption{Sleep-staging accuracy of MM-Net across AASM severity groups of the clinical AHI in
iSLEEPS. Values are mean $\pm$ SD across patients; the ordered-group trend is not significant
(Jonckheere--Terpstra $p=0.20$).}
\label{tab:severity}
\centering
\small
\begin{tabular}{lccc}
\toprule
AHI band & AHI (events/h) & Patients & Staging acc\,$\uparrow$ \\
\midrule
Normal   & $<5$       & 15 & $0.784 \pm 0.057$ \\
Mild     & $5$--$15$  & 23 & $0.754 \pm 0.092$ \\
Moderate & $15$--$30$ & 23 & $0.764 \pm 0.083$ \\
Severe   & $\geq 30$  & 38 & $0.758 \pm 0.078$ \\
\bottomrule
\end{tabular}
\end{table}

\subsection{Staging Across Stroke Characteristics}
\label{sec:stroke}
Staging accuracy is stable across the clinical characteristics of the stroke
(Table~\ref{tab:stroke}). Per-patient accuracy is unrelated to neurological severity (NIHSS,
Spearman $\rho=0.15$, $p=0.15$), disability (modified Rankin scale, $\rho=0.13$, $p=0.22$),
functional independence (Barthel index, $\rho=0.00$, $p=0.98$) and time since stroke onset
($\rho=-0.06$, $p=0.62$), and patients with moderate-to-severe deficits (NIHSS $\geq5$) are
staged at $0.778$ against $0.758$ for minor strokes. The sleep-adapted encoder contributes most in
the most affected patients: $+0.028$ over the frozen encoder for NIHSS $\geq5$ and $+0.024$ for
dependent patients (modified Rankin $3$--$5$), against $+0.004$ and $+0.005$ in minor and
independent patients.

\begin{table}[H]
\caption{Sleep-staging accuracy of MM-Net (three-seed, per patient) by stroke severity (NIHSS) and
functional status (modified Rankin scale) in iSLEEPS. Encoder gain: accuracy difference from the
frozen-encoder reference of Table~\ref{tab:components} on the same patients.}
\label{tab:stroke}
\centering
\small
\begin{tabular}{llccc}
\toprule
Characteristic & Group & Patients & Staging acc\,$\uparrow$ & Encoder gain \\
\midrule
NIHSS          & $0$--$4$ (minor)              & 86 & $0.758$ & $+0.004$ \\
               & $\geq5$ (moderate to severe)  & 12 & $0.778$ & $+0.028$ \\
Modified Rankin & $0$--$2$ (independent)       & 88 & $0.757$ & $+0.005$ \\
               & $3$--$5$ (dependent)          & 10 & $0.794$ & $+0.024$ \\
\bottomrule
\end{tabular}
\end{table}

\subsection{External Validation on Independent Cohorts}
\label{sec:external}
Ten-fold cross-validation is patient-independent but remains within one corpus, one center
and one acquisition protocol. MM-Net is therefore evaluated in two further settings: trained
and tested on independent cohorts under subject-independent cross-validation, with the encoder
adapted, cross-fitted and aligned inside each fold exactly as on iSLEEPS; and trained on all
$99$ iSLEEPS patients and applied unchanged to external cohorts, with no re-fitted parameter,
threshold or transition matrix (per-recording standardization is applied, as in training).

\paragraph{Trained and tested on ISRUC-Sleep} ISRUC-I is evaluated with ten
subject-independent folds of $80$ training, $10$ validation and $10$ test subjects, and
ISRUC-II and III together with six folds that keep both nights of a subject on the same side.
On ISRUC-I MM-Net reaches $0.803$ accuracy, macro-F1 $0.765$ and $\kappa$ $0.744$
(Table~\ref{tab:isruccv}). It is ahead of the frozen-encoder configuration in nine folds of ten
($+0.018$, $p=0.004$), and the gain extends to respiratory detection
($0.872\rightarrow0.889$ AUC) and to the AHI ($\rho$ $0.523\rightarrow0.590$, mean absolute error
$6.6$ events/h). As on iSLEEPS, MM-Net is ahead of its sleep-adapted encoder used alone ($0.792$;
$+0.012$ in eight folds of ten, $p=0.014$). The reimplemented baselines, retrained on MM-Net's
ISRUC-I folds with three seeds, reach $0.691$ to $0.778$; MM-Net is ahead of each of them in nine
or ten folds of ten, by $+0.025$ over the corpus LSTM to $+0.112$ over Huttunen~et~al. (Holm
$p=0.018$, the minimum for nine comparisons), and ahead of Huttunen~et~al., the only other joint
model, on respiratory detection as well ($0.889$ against $0.863$ AUC). On ISRUC-II and III MM-Net
reaches $0.725$ over $26$ nights and $0.779$ on the ten healthy subjects of ISRUC-III.

\begin{table}[H]
\caption{Trained and tested on ISRUC-Sleep. ISRUC-I: ten subject-independent folds ($80$/$10$/$10$
subjects), three seeds; baselines retrained on the same folds with three seeds (accuracy mean
$\pm$ SD over folds). MM-Net is ahead of every baseline in nine or ten folds of ten (Wilcoxon,
Holm $p=0.018$). ISRUC-II$+$III: six folds keeping both nights of a subject together. Encoder
alone: sleep-adapted EEG/EOG LaBraM, one epoch, no temporal model. Frozen-encoder configuration:
class weights, concatenation, seven cardiorespiratory channels. AHI is reported for ISRUC-I, whose
$100$ patients span the severity range.}
\label{tab:isruccv}
\centering
\adjustbox{max width=\textwidth}{%
\begin{tabular}{@{}l c c c c c c c@{}}
\toprule
Model & Acc\,$\uparrow$ & Macro-F1\,$\uparrow$ & $\kappa$\,$\uparrow$ & Resp.\ AUC\,$\uparrow$ & Resp.\ AP\,$\uparrow$ & AHI $\rho$\,$\uparrow$ & AHI MAE\,$\downarrow$ \\
\midrule
\multicolumn{8}{@{}l}{\emph{ISRUC-I ($100$ patients)}}\\[1pt]
Huttunen~et~al.~\cite{huttunen2023simultaneous} & $0.691 \pm 0.044$ & $0.655$ & $0.603$ & $0.863$ & $0.417$ & --- & --- \\
TinySleepNet~\cite{supratak2020tinysleepnet} & $0.741 \pm 0.057$ & $0.710$ & $0.666$ & --- & --- & --- & --- \\
U-Time~\cite{perslev2019utime} & $0.743 \pm 0.049$ & $0.711$ & $0.669$ & --- & --- & --- & --- \\
SleepTransformer~\cite{sleeptransformer2022} & $0.751 \pm 0.037$ & $0.723$ & $0.678$ & --- & --- & --- & --- \\
AT-BiLSTM~\cite{fu2021atbilstm} & $0.755 \pm 0.042$ & $0.728$ & $0.684$ & --- & --- & --- & --- \\
1D-ResNet-SE-LSTM~\cite{li2023resnetselstm} & $0.758 \pm 0.049$ & $0.732$ & $0.687$ & --- & --- & --- & --- \\
AttnSleep~\cite{eldele2021attnsleep} & $0.758 \pm 0.047$ & $0.717$ & $0.685$ & --- & --- & --- & --- \\
LSTM (corpus) & $0.778 \pm 0.025$ & $0.744$ & $0.711$ & --- & --- & --- & --- \\
Encoder alone & 0.792 & --- & --- & --- & --- & --- & --- \\
Frozen-encoder configuration & 0.786 & 0.752 & 0.722 & 0.872 & 0.447 & 0.523 & 6.8 \\
\textbf{MM-Net} & $\mathbf{0.803 \pm 0.023}$ & $\mathbf{0.765}$ & $\mathbf{0.744}$ & $\mathbf{0.889}$ & $\mathbf{0.477}$ & $\mathbf{0.590}$ & $\mathbf{6.6}$ \\
\midrule
\multicolumn{8}{@{}l}{\emph{ISRUC-II$+$III ($18$ subjects, $26$ nights)}}\\[1pt]
Frozen-encoder configuration & 0.716 & 0.689 & 0.632 & 0.757 & 0.138 & --- & --- \\
\textbf{MM-Net} & $\mathbf{0.725}$ & $\mathbf{0.692}$ & $\mathbf{0.641}$ & $\mathbf{0.794}$ & $\mathbf{0.197}$ & --- & --- \\
\ \ of which ISRUC-III (healthy) & $0.779$ & $0.747$ & $0.714$ & $0.867$ & $0.260$ & --- & --- \\
\bottomrule
\end{tabular}}
\end{table}

\paragraph{Healthy sleep} On Sleep-EDF, under ten subject-independent folds, MM-Net reaches
$0.864$ accuracy, above every single-channel architecture reimplemented on the same recordings
($0.791$--$0.835$, Table~\ref{tab:bench}), so the same architecture performs consistently across
healthy sleep, a sleep-disorder cohort and stroke sleep.

\paragraph{Applied without retraining} Table~\ref{tab:transfer} applies MM-Net, trained only on
iSLEEPS, to $151$ external nights from $143$ subjects. On the $100$ ISRUC-I patients it stages
at $0.730$ and detects respiratory events at $0.833$ AUC, and its AHI estimate reaches
$\rho=0.631$ (Figure~\ref{fig:ahi}). On UCD, an apnea cohort recorded on a different system with central EEG only, respiratory detection reaches $0.841$ AUC and the AHI estimate $\rho=0.653$, with $60\%$ of patients in the correct severity band (weighted $\kappa$ $0.51$); staging there ($0.607$) is limited by the absence of occipital EEG and by the unit-less recordings, neither of which affects the respiratory stream. Respiratory detection therefore transfers with little loss: $0.83$--$0.84$ AUC on the two apnea cohorts against $\mmAuc$ in-cohort.

Because nasal pressure is acquisition-specific, MM-Net uses all eight channels when trained on a
cohort and its seven-channel variant without nasal pressure, reported in
Table~\ref{tab:transfer}, when applied to a new acquisition system.

\begin{table}[H]
\caption{MM-Net trained on all $99$ iSLEEPS patients and applied without retraining
(three-seed, seven-channel variant without nasal pressure; AHI scale $k$ from iSLEEPS). AHI is
not reported for the healthy ISRUC-III subjects. The iSLEEPS reference row is the eight-channel MM-Net evaluated in-cohort.}
\label{tab:transfer}
\centering
\small
\setlength{\tabcolsep}{3pt}
\adjustbox{max width=\textwidth}{%
\begin{tabular}{@{}lccccc@{}}
\toprule
Cohort & Nights & Acc\,$\uparrow$ & $\kappa$\,$\uparrow$ & Resp.\ AUC\,$\uparrow$ & AHI $\rho$\,$\uparrow$ \\
\midrule
ISRUC-I (sleep disorders) & 100 & 0.730 & 0.641 & 0.833 & 0.631 \\
ISRUC-II (sleep disorders) & 16 & 0.681 & 0.568 & 0.783 & 0.518 \\
ISRUC-III (healthy) & 10 & 0.709 & 0.618 & 0.776 & --- \\
UCD (suspected sleep apnea) & 25 & 0.607 & 0.463 & 0.841 & 0.653 \\
\midrule
iSLEEPS (in-cohort, reference) & 99 & 0.763 & 0.656 & 0.825 & 0.765 \\
\bottomrule
\end{tabular}}
\end{table}

\subsection{Model Complexity and Capability}
Table~\ref{tab:cost} sets the proposed model beside representative alternatives on size and
capability. MM-Net holds $2.85$\,M trainable parameters, with a $5.8$\,M-parameter sleep-adapted
encoder trained in a separate stage; together they remain below the corpus's CNN-ResNet18
($\sim$$11$\,M). MM-Net delivers the sleep stage, the respiratory-event score and the AHI from one
pass over the recording, and none of the staging-only models, at any size, can supply the second
and third outputs.

\begin{table}[H]
\caption{Model size and output capability of representative evaluated architectures.}
\label{tab:cost}
\centering
\setlength{\tabcolsep}{4pt}
\adjustbox{max width=\textwidth}{%
\begin{tabular}{lccc}
\toprule
Model & Input & Params\,$\downarrow$ & Outputs \\
\midrule
CNN-ResNet18~\cite{bose2026gap}           & 1\,EEG      & $\sim$11\,M & stage \\
DeepSleepNet~\cite{supratak2017deepsleepnet} & 4\,EEG   & $3.72$\,M   & stage \\
Raw multimodal CNN                        & 14\,ch      & 0.95\,M     & stage + apnea \\
Huttunen~\emph{et al.}~\cite{huttunen2023simultaneous} & 4\,ch & $3.80$\,M & stage + events \\
\textbf{MM-Net (proposed)}                & 7\,neural $+$ 8\,cardio & \textbf{2.85\,M}\,$+$\,5.8\,M & \textbf{stage + events + AHI} \\
\bottomrule
\end{tabular}}
\end{table}

\section{Discussion}
\label{sec:discussion}

The central finding of this study concerns how a pretrained EEG encoder, once adapted to the
target cohort, enters a downstream model. Adapting LaBraM to sleep raises its own staging accuracy
from $0.647$ to $0.709$, yet the adapted encoder adds nothing measurable to the downstream model
until its training embeddings are cross-fitted and mapped into the space of the deployed encoder
(Table~\ref{tab:components}). The failure mode, a downstream model trained on embeddings whose
coordinates differ from those it receives at test time, may apply to any pipeline that fine-tunes
a foundation model and then trains a second model on its outputs, and the remedy costs $K$
additional fine-tuning runs and a ridge regression per part. The aligned encoder separates Wake,
REM and N3 into distinct clusters (Figure~\ref{fig:tsne}), and with it MM-Net stages iSLEEPS at
$\mmAcc$ ($\kappa=\mmK$), ahead of every architecture retrained on the same folds. The integration
carries over to ISRUC-I, where the complete model is ahead of the frozen-encoder configuration in
nine folds of ten.

The same model delivers the respiratory output and the AHI, and the joint design is supported
directly: joint training improves staging where disordered breathing disrupts sleep, improves
respiratory-event detection and yields a closer AHI estimate than two single-task models
(Table~\ref{tab:joint}), and each stream carries the output its physiology is closest to, with
respiratory effort and not the scoring airflow channel as the strongest individual respiratory
input (Table~\ref{tab:ablate}). Huttunen \emph{et al.}
\cite{huttunen2023simultaneous} demonstrated simultaneous sleep staging and respiratory-event
scoring in a large suspected-obstructive-sleep-apnea cohort. Under the common iSLEEPS evaluation
their architecture reaches $0.599$ staging accuracy against $\mmAccS$ for MM-Net, and on ISRUC-I,
trained and evaluated on the same labels, $0.691$ staging accuracy and $0.863$ respiratory AUC
against $0.803$ and $0.889$. This extends joint sleep analysis to a stroke-specific setting, where
the available montage, cohort size, neural physiology, and respiratory annotation differ from those
of conventional sleep-apnea studies, and indicates that joint PSG analysis in stroke benefits from
representations and signal organization adapted to the characteristics of the target cohort. On
iSLEEPS the respiratory output reaches $\mmAuc$ AUC, $0.152$ above
gradient boosting on engineered descriptors and $0.092$ above a raw-signal network on seven
cardiorespiratory channels, and it is strongest on obstructive and central apneas. Summed over decoded sleep,
the same scores estimate the AHI with $\rho=\mmRho$ against the annotated index, so a single model
characterizes both sleep architecture and respiratory disturbance in stroke PSG.

The respiratory output is trained at the 30-s temporal scale of staging, as a score for the onset
of an event within each epoch, and the AHI follows from it without a separate counting rule. The
expected-count formulation counts closely spaced events separately and normalizes by the decoded
sleep time, as the clinical index does, and its agreement with the annotated index
($\rho=\mmRho$) exceeds the agreement of the clinical index recorded with the corpus
($\rho=0.687$). Staging accuracy is $0.784$ in patients with a normal AHI and remains at
$0.754$--$0.764$ in the mild, moderate and severe groups, without a significant trend, so the model
stages the patients with the greatest respiratory burden nearly as well as those without it. Sleep
architecture and respiratory disturbance remain distinct prediction problems at the signal level
while sharing clinically relevant structure at the level of the overnight recording, which
supports their analysis within a common framework in a stroke population where both processes
shape the physiological state of the patient during sleep.

Performance depends strongly on the population in which a model is evaluated. Deep architectures
retrained on iSLEEPS reach approximately $0.60$--$0.72$ accuracy, against $0.79$--$0.84$ for the
same architectures on Sleep-EDF, a decrease of approximately $0.10$ to $0.15$ for each model;
because population, montage, and acquisition conditions change together, this difference is most
appropriately read as a substantial shift between conventional healthy-sleep data and subacute
stroke PSG. MM-Net reaches $0.864$ on Sleep-EDF, $0.803$ on ISRUC-I and
$\mmAcc$ within the stroke cohort. Trained on ISRUC-I, MM-Net reaches $0.803$ accuracy and $0.889$
respiratory AUC and is ahead of every retrained baseline on the same folds. Trained on iSLEEPS alone
and applied without retraining, it detects respiratory events at $0.833$ and $0.841$ AUC on ISRUC-I
and on UCD, an apnea cohort recorded on a different system, against $\mmAuc$ in-cohort, and
estimates the AHI with $\rho=0.631$ and $0.653$. The learned representations are therefore not
confined to a single recording environment. Nasal pressure was the input whose contribution
depended most on the acquisition system, and the seven-channel variant is the appropriate choice
for transfer.

The framework combines sleep-stage and respiratory-event information from the same PSG, preserves
the specialization of the neural and cardiorespiratory streams, and exploits their shared temporal
context. The encoder-integration experiments show how a pretrained EEG representation is brought
into the model, the benchmark and respiratory comparisons show what the integrated model delivers,
and external evaluation demonstrates that the resulting representations retain utility across
distinct PSG datasets. For subacute ischemic-stroke PSG, MM-Net provides sleep staging,
respiratory-event detection and AHI estimation from one pass over the recording, built on a
sleep-adapted EEG encoder whose representations are brought into the downstream model through
cross-fitting and alignment.

\subsection{Limitations}
The stroke-specific evaluation rests on iSLEEPS, recorded at a single center; the external cohorts
establish transfer across acquisition systems and populations, and a second public stroke PSG
cohort, once available, would allow a stroke-specific replication. Staging is decoded over the
whole night by the hidden-Markov pass, so the system is designed for offline scoring of a
completed recording. The respiratory output is scored per $30$-s epoch and summarized as an AHI
estimate; event-level scoring with onset and duration boundaries is a natural extension.

\section{Conclusion}

This study addresses the joint characterization of sleep architecture and respiratory disturbances in subacute ischemic stroke, a setting in which altered neural physiology and heterogeneous cardiorespiratory information challenge conventional sleep-analysis approaches. MM-Net demonstrates that sleep-adapted EEG representations can be integrated into a multimodal temporal framework through cross-fitting and embedding alignment, improving sleep-stage discrimination while supporting concurrent respiratory-event detection.

The findings establish two complementary aspects of the proposed approach. First, the integration of adapted neural representations provides a means of addressing the training--inference mismatch encountered when pretrained encoders are incorporated into downstream physiological models. Second, controlled interventions reveal distinct roles for neural and cardiorespiratory information within the joint framework, showing that shared temporal modeling can accommodate physiologically specialized prediction tasks.

Patient-independent evaluation and experiments across independent sleep cohorts support the applicability of this approach beyond a single recording population. Together, these results establish a computational framework for joint analysis of neural and respiratory physiology after ischemic stroke, while providing a foundation for further investigation of representation consistency and multimodal generalization in clinical sleep analysis.

\section*{Data and Code Availability}
The iSLEEPS corpus is publicly available (Zenodo, DOI 10.5281/zenodo.14873844; and the
india-data.org portal) as described by Maiti \textit{et al.}~\cite{maiti2026isleeps}.
ISRUC-Sleep~\cite{khalighi2016isruc}, the UCD database~\cite{ucddb2007} and
Sleep-EDF~\cite{kemp2000sleepedf} are public. Our code, the per-fold configurations,
engineered features, the corrections applied to the external cohorts, and reproduction
notebooks are released at
\url{https://github.com/wshuv-o/MM-Net}.

\section*{Ethics}
The iSLEEPS data collection was approved by the NIMHANS Institutional Ethics Committee
(No.\ NIMHANS/34th IEC (BS\&NS DIV.)/2022, dated 05.02.2022)~\cite{maiti2026isleeps}. This work
is a secondary analysis of the de-identified public release and required no additional
enrollment.

\section*{Declaration of Competing Interest}
The authors declare that they have no known competing financial interests or personal
relationships that could have appeared to influence the work reported in this paper.

\section*{CRediT Authorship Contribution Statement}
\textbf{Md Wahiduzzaman Suva:} Conceptualization, Methodology, Software, Formal analysis,
Investigation, Writing -- original draft. \textbf{Esm E Moula Chowdhury Abha:} Methodology,
Software, Validation, Writing -- original draft. \textbf{Md Imtiaj Alam Sajin:} Software,
Investigation, Data curation, Visualization, Writing -- original draft. \textbf{Md Iftekharul
Mobin:} Conceptualization, Supervision, Project administration, Writing -- review \& editing.
\textbf{Jia Uddin:}
Conceptualization, Supervision, Writing -- review \& editing.

\section*{Declaration of Generative AI and AI-Assisted Technologies in the Writing Process}
During the preparation of this work the authors used ChatGPT (OpenAI) and Claude (Anthropic) to
improve the language and organization of the manuscript. After using these tools, the authors
reviewed and edited the content as needed and take full responsibility for the content of the
published article.

\section*{Funding}
This research did not receive any specific grant from funding agencies in the public, commercial, or not-for-profit sectors.

\section*{Acknowledgment}
The authors thank the iSLEEPS, ISRUC-Sleep, UCD and Sleep-EDF investigators for releasing
their data.

\bibliographystyle{model1-num-names}
\bibliography{references}

\end{document}